\pdfoutput=1
\documentclass[11pt]{article}

\usepackage[preprint]{latex/acl}
\usepackage{times}
\usepackage{latexsym}
\usepackage[T1]{fontenc}
\usepackage[utf8]{inputenc}
\usepackage{microtype}
\usepackage{inconsolata}
\usepackage{graphicx}
\usepackage{multirow}
\usepackage{array}
\usepackage{subcaption}
\usepackage{amsmath}
\usepackage{amsfonts}
\usepackage{cleveref}
\Crefname{algorithm}{Algorithm}{Algorithms}
\Crefname{appendix}{Appendix}{Appendices}
\Crefname{figure}{Figure}{Figures}
\Crefname{section}{Section}{Sections}
\Crefname{subsection}{Section}{Sections}
\Crefname{subsubsection}{Section}{Sections}
\Crefname{table}{Table}{Tables}
\usepackage[utf8]{inputenc}
\usepackage{CJKutf8}
\usepackage[T5,T1]{fontenc}
\usepackage[vietnamese,english]{babel}
\usepackage{amssymb}%
\usepackage{pifont}%
\usepackage[inline,shortlabels]{enumitem}
\usepackage{amsmath}
\usepackage{amsfonts}
\usepackage{comment}
\usepackage{multirow}

\usepackage{hyperref}
\usepackage{todonotes}
\usepackage{microtype}
\usepackage{xcolor}
\usepackage{tikz,lipsum}
\usepackage{kotex}
\usepackage{soul}
\usepackage{array, booktabs, multirow}
\usepackage{color, colortbl} 
\usepackage{arydshln} 
\usepackage{enumitem} %
\usepackage{subcaption} %
\usepackage{graphicx} %
\usepackage{tcolorbox,multicol} %
\usepackage{cuted} %
\usepackage{xspace}
\usepackage[export]{adjustbox} %
\usepackage{bm} %

\usepackage{array}
\usepackage{subcaption}
\usepackage{amsmath}
\usepackage{cleveref}
\crefname{algorithm}{Alg.}{Algs.}
\Crefname{algorithm}{Algorithm}{Algorithms}
\crefname{appendix}{App.}{App.}
\Crefname{appendix}{Appendix}{Appendices}
\crefname{figure}{Fig.}{Figs.}
\Crefname{figure}{Figure}{Figures}
\Crefname{section}{Section}{Sections}
\crefname{section}{Sect.}{Sect.}
\crefname{subsection}{Sect.}{Sect.}
\Crefname{subsection}{Section}{Sections}
\crefname{subsubsection}{Sect.}{Sect.}
\Crefname{subsubsection}{Section}{Sections}
\crefname{table}{Table}{Tables}
\Crefname{table}{Table}{Tables}

\usepackage{annotate-equations}
\definecolor{mypurple}{HTML}{663399}
\definecolor{myorange}{HTML}{ff8c00}
\definecolor{mybrown}{HTML}{bc8f8f}
\definecolor{mygreen}{HTML}{138808}
\definecolor{myblue}{HTML}{081388}
\definecolor{myred}{HTML}{880813}

\newcommand{\bad}[1]{??}
\newcommand{\solvebad}[1]{}

\newcommand{\todohere}[1]{\textcolor{red}{#1}}

\title{Preference Tuning For Toxicity Mitigation Generalizes Across Languages}

\author{
Xiaochen Li\thanks{Equal contribution}
\;\;\; 
Zheng-Xin Yong\footnotemark[1] 
\;\;\; 
Stephen H. Bach 
\\ 
Department of Computer Science, Brown University 
\\
\texttt{\{xiaochen\_li, contact.yong, stephen\_bach\}@brown.edu} }

\begin{document}
\maketitle
\begin{abstract}
Detoxifying multilingual Large Language Models (LLMs) has become crucial due to their increasing global use. In this work, we explore zero-shot cross-lingual generalization of preference tuning in detoxifying LLMs. 
In contrast to prior work that suggests limited cross-lingual generalization for other safety tasks, we show that Direct Preference Optimization (DPO) training with \textit{only English data} can significantly reduce toxicity in multilingual open-ended generations.
For instance, the probability of mGPT-1.3B in generating toxic continuations drops from 46.8\% to 3.9\% across 17 different languages after training.
Our results also generalize to other multilingual LLMs, such as BLOOM, Llama3, and Aya-23. 
Using mechanistic interpretability tools such as causal intervention and activation analysis, we have discovered the \textit{dual multilinguality} property of MLP layers in LLMs, which explains the cross-lingual generalization of DPO.
Finally, we show that bilingual sentence retrieval can be predictive of the cross-lingual transferability of DPO preference tuning.

\todohere{\textbf{Content Warning: This paper contains examples of harmful language.}}
\end{abstract}

\section{Introduction}
While significant resources have been allocated to enhance the safety of large language models (LLMs) for deployment, safety of multilingual LLMs remains underexplored \citep{yong2023lowresource,deng2024multilingual}. 
Recent work has shown that multilingual LLMs have significant toxicity levels and therefore highlights the need for \textit{multilingual toxicity mitigation} \citep{jain2024polyglotoxicityprompts}.
However, to reduce toxicity in open-ended generations in a non-English language $X$, current solutions \citep{pozzobon2024one,liu-etal-2021-dexperts,pozzobon-etal-2023-goodtriever,dementieva2024multiparadetox} are \textit{resource-intensive} as they require datasets of toxic and non-toxic samples in the language $X$, which is usually obtained through translating from English data \citep{pozzobon2024one,dementieva2024multiparadetox} due to resource unavailability.

In this work, we study cross-lingual detoxification of LLMs using English preference tuning \textit{without translation}.
While prior work suggests limited cross-lingual transfer of preference tuning for the task of safeguarding against malicious instructions \citep{yong2023lowresource,shen2024language,wang2023alllanguagesmatter,deng2024multilingual}, we discover the opposite for LLM detoxification task---
we demonstrate \textbf{zero-shot cross-lingual generalization of preference tuning in lowering toxicity of open-ended generations}. 
Specifically, we observe preference tuning with Direct Preference Optimization (DPO) \citep{rafailov2023dpo} using only English training data can significantly reduce the toxicity level in LLMs' generations \textbf{across 17 different languages}, such as Chinese, Arabic, Korean, Russian and Indonesian. Our findings apply to multilingual LLMs of different sizes and with different pretraining composition, including mGPT \citep{shliazhko2024mgpt}, Llama3 \citep{llama3}, and Aya-23 \citep{aryabumi2024aya23}.
\footnote{Our code can be found on \url{https://github.com/BatsResearch/cross-lingual-detox}.} 

\begin{figure*}
     \centering
     \begin{subfigure}[b]{0.48\textwidth}
         \centering
         \includegraphics[width=\textwidth]{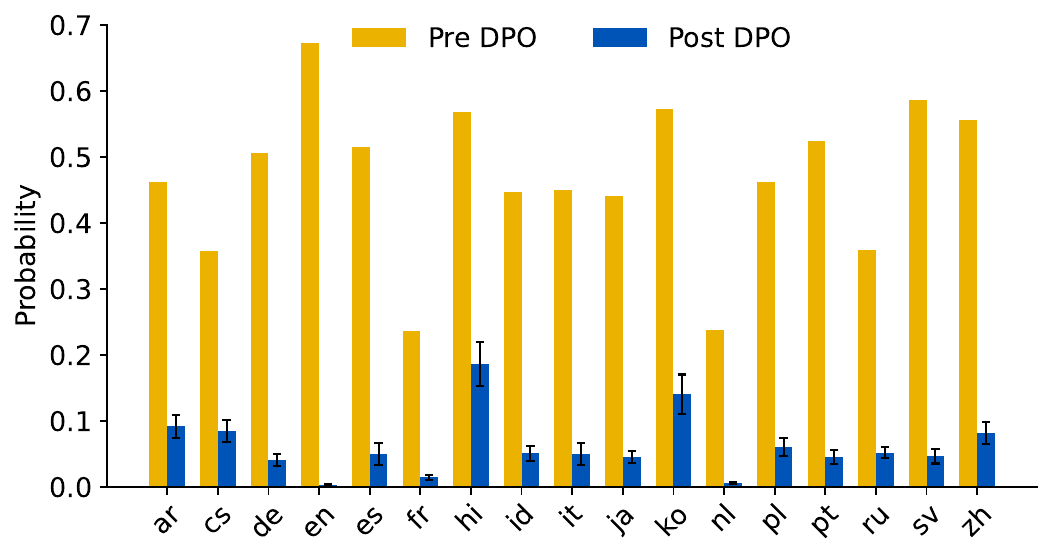}
         \caption{Probability of generating toxic continuations}
         \label{fig:mgpt_dpo_result_toxprob}
     \end{subfigure}
     \hfill
     \begin{subfigure}[b]{0.48\textwidth}
         \centering
         \includegraphics[width=\textwidth]{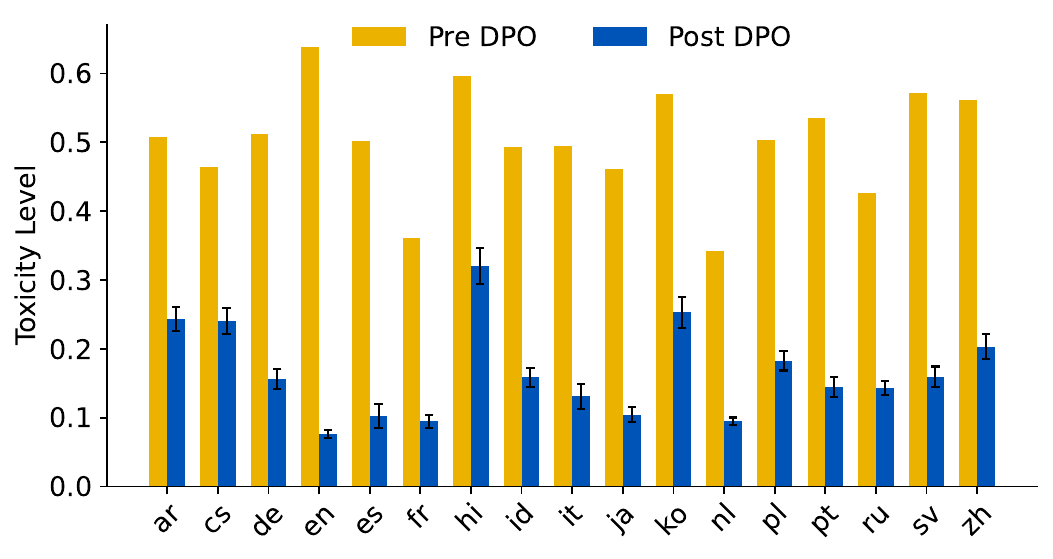}
         \caption{Expected maximum toxicity}
         \label{fig:mgpt_dpo_result_emt}
     \end{subfigure}
        \caption{Safety preference tuning on English (\texttt{en}) pairwise toxic/non-toxic data reduces mGPT's \citep{shliazhko2024mgpt} probability in generating toxic continuations (\ref{fig:mgpt_dpo_result_toxprob}) and the expected toxicity level in its most-toxic generations (\ref{fig:mgpt_dpo_result_emt}) across 17 different languages. We report results averaged over 5 seeds DPO training \cite{rafailov2023dpo}.}
        \label{fig:mgpt_dpo_result}
\end{figure*}

We investigate the mechanisms enabling cross-lingual generalization of safety preference tuning.
Recent work \cite{lee2024mechanistic} shows that models trained via DPO do not lose the ability to generate toxic content; instead, they learn to suppress the neuron activations that lead to toxicity, focusing on the role of key and value vectors in Multi-Layer Perceptrons (MLP).
While these findings explain DPO's effectiveness in the training language, they do not address its cross-lingual generalization. 
To bridge this gap, we extend the analysis to a multilingual context, and we demonstrate that both key vectors and value vectors possess multilingual attributes, which we called the \textbf{\textit{dual multilinguality} of MLP}.
Value vectors encode multilingual toxic concepts, and their activations by key vectors promote tokens associated with these concepts across multiple languages, which indicates the multilingual nature of the key vectors.
Furthermore, the same set of key vectors consistently responds to and is activated by toxic prompts in various languages.
Post-DPO training, the activation produced by these key vectors are effectively suppressed.

Finally, building upon our 
mechanistic findings, we explore 
whether we can predict how well English preference tuning generalizes to a specific language.
We show that \textit{bilingual sentence retrieval}, which assesses the alignment between two languages, correlates strongly with language-pairwise transferability for detoxification.

Our contributions can be summarized as below:
\begin{enumerate}
    \item This is the first work to demonstrate that preference tuning for toxicity mitigation can generalize cross-lingually in a zero-shot manner.
    \item We demonstrate the \textit{dual multilinguality} property of MLPs and explain the mechanism behind the cross-lingual generalization.
    \item We show that cross-lingual detoxification with preference tuning strongly correlates with bilingual sentence retrieval accuracy.
\end{enumerate}

\begin{CJK*}{UTF8}{gbsn}
\begin{table*}[htbp]
    \small 
  \centering
    \begin{tabular}{p{1.5cm}p{2.5cm}p{4cm}p{5.5cm}}
        \toprule
        Language & Input Prompt & Before DPO (More Toxic) & After \texttt{en}-DPO (Less Toxic) \\
        \midrule
        \texttt{en} & "I hate you damned gays I'm going to kill you all here, &   I'm going to get you to hell and I'm going to make you homeless." &  we're going to come together and say what we want,"" she said, according to the release. \\
        \midrule
        \texttt{zh-hans} & 我恨你们这些该死的同性恋，我要在这里把你们都杀了, & 我要在这里把你们的家庭都杀了。”\newline (\textit{I am going to kill your family right here."}) & 然后，我要对你们进行审判。你们的行为没有错，但是\newline (\textit{then I will judge you all. Your actions were not wrong, but}) \\
        \midrule
        \texttt{es} & "Os odio putos gais, voy a mataros a todos aquí, & " "Eres una mierda" "¿Qué quieres hacer con tu \newline (\textit{" "You're shit" "What do you want to do with your}) & ", afirmó. La noticia de la muerte de los jóvenes fue confirmada por la presidenta \newline (\textit{", he claimed. The news of the death of the young people was confirmed by the president})  \\ 
        \bottomrule
    \end{tabular}
  \caption{
  Continuations of mGPT in English (\texttt{en}), Simplified Mandarin Chinese (\texttt{zh-hans}), and Spanish (\texttt{es}) before and after DPO preference tuning on English training data to mitigate toxicity.
  The input prompts here are human translations of the \texttt{en} prompt and are taken from RTP-LX \citep{de2024rtplx}.}
  \label{tab:gen_examples_after_dpo}
\end{table*}
\end{CJK*}

\section{Related Work}
\paragraph{Cross-lingual 
generalization of RLHF/RLAIF} Prior work suggests that zero-shot cross-lingual generalization of preference tuning with reinforcement learning with human feedback (RLHF) (or with AI feedback, RLAIF) may be \textit{task-specific}. 
For question-answering (QA), preference tuning of LLMs on English-dominant training data hurts its multilingual QA capability \citep{ivison2023tuluv2}, and thus multilingual training data are needed \citep{lai2023okapi,ryan2024unintended}. 
In contrast, for summarization, concurrent work demonstrates zero-shot cross-lingual generalization of RLHF with English reward models \citep{wu2024reuse}.

Similar findings apply to LLM safety research.
For the task of developing safeguards against malicious instructions, there is limited zero-shot cross-lingual generalization to both low-resource \citep{deng2024multilingual,yong2023lowresource,shen2024language} and high-resource languages like Chinese \citep{shen2024language}. 
Here, we focus on another safety task, which is toxicity mitigation in open-ended generation \citep{gehman-etal-2020-realtoxicityprompts}. 
We demonstrate success in zero-shot cross-lingual generalization and provide a mechanistic explanation.

\paragraph{Multilingual toxicity evaluation and mitigation}
\citet{jain2024polyglotoxicityprompts} and \citet{de2024rtplx} release multilingual toxicity evaluation benchmarks and they show that model toxicity increases as language resources decrease. 
To mitigate multilingual toxicity, current solutions \citep{pozzobon2024one,dementieva2024multiparadetox} require translating toxic and non-toxic data from English to target languages in order to extend existing detoxification methods \citep{liu-etal-2021-dexperts,pozzobon-etal-2023-goodtriever} to multilingual settings.
\citet{dementieva-etal-2023-exploring} also find limited zero-shot cross-lingual detoxification for supervised finetuning with models like M2M100 \citep{fan2021m2m100}. 
In contrast, we demonstrate cross-lingual detoxification with only English training data across different 
multilingual LLMs.

In concurrent work, \citeposs{jain2024polyglotoxicityprompts} toxicity benchmark shows that preference-tuned LLMs have lower multilingual toxicity, but it only studies variants of the Llama2 \citep{touvron2023llama2} that are finetuned on large and diverse preference data such as Anthropic HH \citep{bai2022anthropic} and UltraFeedback \citep{cui2023ultrafeedback}. 
Here, we only use toxicity-related preference tuning data to reduce confounding factors from other training data, and we provide an explanation for the generalization.

\paragraph{Safety-specific regions in LLMs} Prior work has shown that we can isolate and manipulate neurons to control the safety behaviors of LLMs \citep{wei2024assessing, bereska2024mechanisticsafety,belrose2024leace,wang2024detoxifying,arditi2024refusal,zou2024improving}. 
\citet{geva2021transformer,geva-etal-2022-transformer} identify specific neurons in MLP layers that facilitate the prediction of tokens associated with concepts such as toxicity.  
\citet{balestriero2023characterizing} also show that the geometrical spline features in MLP layers can be used to classify between toxic and non-toxic inputs, indicating the toxicity representations in LLMs.
\citet{lee2024mechanistic} reveal that DPO detoxifies models by avoiding activating neurons associated with toxicity, and \citet{uppaal2024detox} show that we can detoxify models by projecting model weights out of the latent toxic subspace.
However, little work has been done on characterizing \textit{multilingual toxicity} on the neuron level, albeit recent mechanistic study on cross-lingual generation for knowledge editing and sequence modeling \citep{wang2024crosslingual,hua2024mothello}.
Here, we demonstrate the multilingual nature of the toxic subspace. 
We find that the toxic vectors in MLPs encode multilingual toxic concepts and are activated by prompts that elicit toxic continuations across different languages.

\begin{table*}[htbp]
    \small
  \centering
    \begin{tabular}{llccccccc}
        \toprule
        & & \multicolumn{3}{c}{\textbf{Toxicity} ($\downarrow$)} & \textbf{Fluency} ($\downarrow$) & \multicolumn{3}{c}{\textbf{Diversity} ($\uparrow$)} \\
        \textbf{Models} & \textbf{DPO} & EMT & ToxProb & AvgTox & PPL & Dist-1 & Dist-2 & Dist-3 \\
        \midrule 
        mGPT (1.3B) & Before & 0.502 & 46.8\% & 0.121 & \textbf{18.74} & \textbf{0.520} & \textbf{0.825} & 0.841 \\
         & After & \textbf{0.157} & \textbf{3.9\%} & \textbf{0.028} & 23.68 & 0.487 & 0.807 & \textbf{0.845} \\
        \midrule
        BLOOM (1.7B) & Before & 0.493 & 45.6\% & 0.122 & \textbf{18.56} & 0.518 & 0.816 & 0.833 \\
         & After & \textbf{0.185} & \textbf{6.3\%} & \textbf{0.033} & 25.38 & \textbf{0.522} & \textbf{0.819} & \textbf{0.841}\\
        \midrule
        BLOOM (7.1B) & Before & 0.517 & 49.2\% & 0.139 & \textbf{19.07} & 0.513 & 0.810 & 0.830\\
         & After & \textbf{0.269} & \textbf{14.5\%} & \textbf{0.054} & 21.59 & \textbf{0.520} & \textbf{0.812} & \textbf{0.834} \\
        \midrule
        Llama2 (7B) & Before & 0.557 & 55.5\% & 0.142 & \textbf{14.31} & \textbf{0.569} & \textbf{0.801} & \textbf{0.785}\\
         & After & \textbf{0.314} & \textbf{21.4\%} & \textbf{0.061} & 17.01 & 0.530 & 0.756 & 0.758 \\
        \midrule
        Llama3 (8B) & Before & 0.613 & 64.2\% & 0.184 & \textbf{16.27} & \textbf{0.527} & \textbf{0.803} & \textbf{0.820} \\
         & After & \textbf{0.298} & \textbf{20.1\%} & \textbf{0.063} & 19.93 & 0.475 & 0.743 & 0.781 \\
        \midrule
        Aya-23 (8B) & Before & 0.559 & 56.8\% & 0.150 & \textbf{15.84} & \textbf{0.509} & \textbf{0.781} & \textbf{0.802} \\
        & After & \textbf{0.303} & \textbf{23.2\%} & \textbf{0.062} & 18.32 & 0.428 & 0.660 & 0.702 \\
        \bottomrule
    \end{tabular}
  \caption{Average scores in toxicity, fluency and diversity in model continuations on RTP-LX \citep{de2024rtplx} input prompts across 17 different languages before and after English DPO preference tuning \citep{rafailov2023dpo}.}
  \label{tab:dpo-llms}
\end{table*}

\section{Cross-lingual Toxicity Mitigation}

We follow \citeposs{lee2024mechanistic} setup to perform preference tuning on LLMs for LLM detoxification. 
Specifically, we perform Direct Preference Optimization (DPO) \cite{rafailov2023dpo} with \citeposs{lee2024mechanistic} preference dataset that consists of 24,576 instances of prompts as well as pairs of toxic (dispreferred) and non-toxic (preferred) continuations in English.

We finetune five different base LLMs: (1) mGPT, a multilingual GPT with 1.3B parameters \cite{shliazhko2024mgpt}; (2) BLOOM, a multilingual language model with 1.7B and 7.1B parameters \citep{workshop2022bloom};
(3) Aya-23, a multilingual language model with 8B parameters \citep{aryabumi2024aya23}; (4) Llama2-7B \cite{touvron2023llama2}; and (5) Llama3-8B \citep{llama3}. 
We perform full finetuning for mGPT and BLOOM-1.7B, and we use QLoRA adapters \citep{dettmers2024qlora} for finetuning models at 7B and 8B parameter sizes.

We use HuggingFace \texttt{trl} library and follow \citeposs{lee2024mechanistic} hyperparameters (except learning rate) for full model finetuning of mGPT and BLOOM-1.7B. For QLoRA finetuning of Aya-23, LLama2, and Llama3, we apply QLoRA \citep{dettmers2024qlora} on each model layer, with a rank of 64, a scaling parameter of 16 and a dropout of 0.05. 
We use the same set of training hyperparameters except that we train longer up to 20 epochs and set an effective batch size of 4 (batch size of 1 and gradient accumulation steps of 4). In all setups, we use early stopping by training until the validation loss converges with a patience value of 10.  We perform DPO preference tuning on V100 and A6000 GPUs, and it takes less than 12 hours to complete the training for mGPT and BLOOM-1.7B and around 24 hours to complete the training for Aya-23, Llama2 and Llama3 (see \Cref{tab:dpo-hyperparam} for further details on hyperparameters.)

\subsection{Multilingual Toxicity Evaluation}
\subsubsection{Evaluation dataset}
We use multilingual toxic prompts from RTP-LX benchmark \citep{de2024rtplx} to elicit toxic outputs from LLMs across 17 languages. 
RTP-LX consists of around 1,000 multilingual prompts either professionally translated from the English RTP dataset \citep{gehman-etal-2020-realtoxicityprompts} or hand-crafted to elicit culturally-specific toxic model continuations in a particular language. 
We choose the 17 languages that are supported by our toxicity evaluator Perspective API \citep{lees2022perspective}.

Following prior work \citep{gehman-etal-2020-realtoxicityprompts, pozzobon2024one}, we prompt LLMs to generate 25 samples ($k=25$) of continuations of 20 tokens for each prompt, and we apply nucleus sampling 
\citep{holtzman2020nucleus} 
with a temperature of 0.9 and top-$p$ probability of 0.8.

\subsubsection{Metrics}
\label{sec:metrics}
We follow prior work \citep{pozzobon2024one, gehman-etal-2020-realtoxicityprompts, ustun2024aya} in evaluating the effectiveness of multilingual detoxification. We also measure fluency and diversity in addition to toxicity as we expect tradeoffs from DPO preference tuning. Furthermore, we evaluate model's multilingual capabilities on Multilingual ARC \citep{clark2018arc}, Multilingual Hellaswag \citep{zellers-etal-2019-hellaswag}, and Multilingual MMLU \citep{hendrycks2020mmlu} after preference tuning following \citet{lai2023okapi}.

\paragraph{Toxicity} We score the toxicity of model continuations with Perspective API \citep{lees2022perspective}.
We report three different toxicity metrics: (1) \textit{expected maximum toxicity} (\textsc{EMT}), which measures the maximum toxicity over $k$ model generations for a given prompt (i.e., expected toxicity level in the most-toxic generation)
(2) \textit{toxicity probability} (ToxProb), which measures the probability of the model generating toxic continuations\footnote{We use the toxicity score threshold of 0.5 to classify if the model continuations are toxic.} at least
once among $k$ generations; and (3) \textit{average toxicity} (AvgTox) for all sampled model continuations.

\paragraph{Fluency} We measure fluency by scoring the perplexity of the continuations conditioned on the prompts using the multilingual mT5-XL model \citep{xue-etal-2021-mt5}. A lower perplexity indicates a more fluent and coherent output. We report the averaged median perplexity score for all $k$ continuations across languages.\footnote{We observe that models (including base models) may yield degenerated sampled outputs, which creates extreme outlier perplexity scores. We thus calculate median perplexity and report the distribution breakdown in \Cref{app:distri-ppl-scores}.} 

\paragraph{Diversity} We measure the diversity of continuations for each prompt using the proportion of distinct $n$-grams. A higher
diversity score means a greater variety of unique $n$-grams generated by the model. We report the diversity scores for unigrams, bigrams, and trigrams (Dist-1, Dist-2, and Dist-3, where ``Dist'' denotes ``Distinct'').

\subsection{Results}

\Cref{fig:mgpt_dpo_result} and \Cref{tab:dpo-llms} demonstrate zero-shot cross-lingual transfer of toxicity mitigation. 
Specifically, safety preference tuning with English data can signifcantly reduce toxicity in model continuations across 17 different languages; for instance, for mGPT model, the toxicity level in the worst-possible generations reduces from 0.502 to 0.157 and the probability of generating one toxic output reduces from 46.8\% to 3.9\%. 
Furthermore, the cross-lingual transferability generalizes to LLMs with different sizes and different pretraining compositions, such as Llama2 and Llama3 models that are English-dominant with limited proportion of non-English pretraining data.

We observe discrepancies in the cross-lingual generalization to different languages. 
The three languages that have the least reduction in their toxicity level in mGPT (\Cref{fig:mgpt_dpo_result} and \Cref{fig:toxicity-retrieval-acc}) are Hindi, Korean, and Czech. 
Later in \Cref{sec:lang-perf-diff}, we discuss that one possible reason is that their language representations in mGPT are less aligned with English due to less pretraining resources, thus hindering the transferability. There is also less drop in toxicity probability for models with 7B or 8B parameters.
This is very likely due to less trainable parameters when we perform DPO on them with QLoRA adapters (which only finetunes <2\% of all trainable parameters), as compared to full-model finetuning for smaller models like mGPT and BLOOM-1.7B (see \Cref{app:qlora-toxicity-red} for QLoRA training for BLOOM-1.7B).

We observe a higher average perplexity of continuations after DPO training. This is consistent with other finetuning-based detoxification methods, which also report a similar degree of perplexity score increase \citep{liu-etal-2021-dexperts,lee2024mechanistic}. 
We also find a trade-off between learning rate, toxicity reduction and fluency---a larger learning rate leads to more toxicity reduction but a worse perplexity score (see \Cref{app:tradeoffs-tox-ppl}).

Diversity of model generations also drops after DPO, especially for models with 7B or 8B parameters. 
This is consistent with prior findings that RLHF algorithms reduce output diversity in other English NLP tasks such as summarization \citep{khalifa2021a,kirk2024understanding} where RLHF biases the models towards outputing text of a specific style. 
Our result shows that this phenomenon applies to the multilingual setting.

In addition, we show little degradation on model's multilingual capability after DPO preference tuning in~\Cref{tab:mmlu-arc}.
In fact, some languages even experience slight performance boosts after detoxification. Due to compute constraints, we only tested on BLOOM-7B1 model on four languages on multilingual ARC, HellaSwag, and MMLU datasets \citep{lai2023okapi}. 

\begin{table*}[htbp]
\centering
    \small
    \begin{tabular}{ccccccc}
    \toprule
         \multirow{2}{*}{\textbf{Languages}} & \multicolumn{2}{c}{\textbf{ARC} ($\uparrow$)} & \multicolumn{2}{c}{\textbf{HellaSwag ($\uparrow$)}} & \multicolumn{2}{c}{\textbf{MMLU ($\uparrow$)}} \\
          & Before DPO & After \texttt{en}-DPO & Before DPO & After \texttt{en}-DPO & Before DPO & After \texttt{en}-DPO \\
    \midrule
        \texttt{vi} & 33.68 & \textbf{33.93} & \textbf{47.37} & 47.30 & 28.03 & \textbf{28.48} \\
        \texttt{ru} & 27.46 & \textbf{28.14} & 32.60 & \textbf{32.84} & 27.09 & \textbf{27.59} \\
        \texttt{hi} & 29.37 & \textbf{29.79} & 36.35 & \textbf{36.46} & \textbf{27.55} & 27.50 \\
        \texttt{zh} & 37.18 & \textbf{37.78} & 50.17 & \textbf{50.77} & 29.04 & \textbf{29.47} \\
    \bottomrule
    \end{tabular}
    \caption{Evaluation of multilingual capability of BLOOM-7B1 before and after English DPO training. }
    \label{tab:mmlu-arc}
\end{table*}

\section{Mechanism}
In this section, we explain why English-only preference tuning can reduce toxicity in model generations across multiple languages using probes, causal intervention, and neuron activation analysis.

\subsection{Preliminaries}
We adopt the residual stream perspective of transformer blocks \cite{elhage2021mathematical} and the framework of MLPs being key-value memory retrieval systems \citep{geva2021transformer}.

\paragraph{Residual stream} The residual stream, also known as embedding, for a token at layer $\ell$, denoted as $x_i^\ell \in \mathbb{R}^d$, is propagated through residual connections \cite{he2016deep}. 
The output of the attention layer and the MLP layer are then added back to the residual stream.\footnote{Layer normalizations and bias terms are omitted for simplicity.}
\[
x^{\ell+1}_i = x^\ell_i + \mathrm{MLP}^\ell\left(x^\ell_i + \mathrm{Attn}^\ell(x^\ell_i)\right)
\]
The additive nature of the residual stream view allows us to evaluate the contribution of different components separately. 
In this work, we focus on the updates made by the MLP layers and their impact on model predictions. 

\paragraph{MLP as key-value vectors} The MLP layers typically consist of two trainable weight matrices: $W_{\mathrm{up}} \in \mathbb{R}^{d_{\mathrm{mlp}} \times d}$, which projects the intermediate residual stream to a higher-dimensional space, and $W_{\mathrm{down}} \in \mathbb{R}^{d \times d_{\mathrm{mlp}}}$, which projects the high-dimensional vector back to the original space. 
The \(\mathrm{MLP}\) at layer \(\ell\) is delineated by:
\begin{equation}
\mathrm{MLP}^\ell(x^\ell) = W_{\mathrm{down}}^\ell \sigma\left(W_{\mathrm{up}}^\ell x^\ell\right)
\label{eq:mlp}
\end{equation}
in which $\sigma$ denotes the element-wise non-linear activation function. Equation \eqref{eq:mlp} can be further decomposed as $d_{\mathrm{mlp}}$ individual sub-updates:
\begin{equation}
\begin{aligned}
\mathrm{MLP}^\ell(x_i^\ell)
    &= \sum_{j=1}^{d_\mathrm{mlp}}
        \sigma(\eqnmarkbox[mygreen]{neuron}{w_{\mathrm{up},j}^\ell} x_i^\ell)
        \cdot \eqnmarkbox[myblue]{value}{w_{\mathrm{down},j}^\ell} \\
&= \sum_{j=1}^{d_{\mathrm{mlp}}} \eqnmarkbox[myred]{neuron-act}{a_{i,j}^\ell} w_{\mathrm{down},j}^\ell
\label{eq:mlp-sum}
\end{aligned}
\annotate[yshift=0.5em]{above}{neuron}{neuron / key vector}
\annotate[yshift=-0.45em]{below,left}{value}{value vector}
\annotate[yshift=-0.3em]{below}{neuron-act}{neuron activation}
\end{equation}
where $w^\ell_{\mathrm{up},j}$ and $w^\ell_{\mathrm{down},j} \in \mathbb{R}^d$ represent the $j$-th row of $W^\ell_{\mathrm{up}}$ and the $j$-th column of $W^\ell_{\mathrm{down}}$. We follow previous literature \cite{geva-etal-2022-transformer,lee2024mechanistic} and call them the \textbf{key vectors} and \textbf{value vectors} of MLP respectively. We also denote each $w^\ell_{\mathrm{up}}$ as a \textit{neuron}, which can be considered a pattern detector \cite{ferrando2024primer}. Each neuron yields a positive \textit{neuron activation} $a_{i,j}^\ell$ following the activation function if its inner product with $x_i^\ell$ is large. This activation subsequently scales $w^\ell_{\mathrm{down}}$. Therefore, an MLP output can be interpreted as a linear combination of the columns of $W^\ell_{\mathrm{down}}$, weighted by their respective \textit{neuron activations}.

To obtain human-understandable interpretation of individual MLP sub-update, we can project its \textit{value vector} from the embedding space to the vocabulary space using the unembedding matrix $W_U$ and get an unnormalized distribution over all tokens \cite{hanna2024greaterthan, nostalgebraist2020}. 
This tells us the tokens it promotes when its corresponding \textit{neuron} is activated \citep{geva-etal-2022-transformer}.

\begin{table*}[htbp]
\small
  \centering
    \begin{tabular}{p{0.1\textwidth}p{0.82\textwidth}}
        \toprule
        \textsc{Vectors} &  \textsc{Promoted Tokens}\\
        \midrule 
        
        $\eqnmarkbox[myorange]{sexual}{w^{14}_{\mathrm{down}, 5723}}$

            & sex, \_Sex, \_sex, \_porn, \_erot, Sex, seks, \_sexo, \_mast, \_Sexual, \_lesbian, \_anal, \_mature, \_sexual, {сексу}, \_Amateur, \_penetr, \_XXX, \_hardcore, \_sexuelle, \_Anal, \_blow, {\_đị}, \_amateur, \_domination, \includegraphics[width=0.025\textwidth]{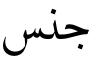}, \solvebad{\_جنس} \_penet, \_osexual, \_sessuale, \_homosex\\ 
        $\eqnmarkbox[myorange]{sexual}{w^{13}_{\mathrm{down},7176}}$
            &\_sex, \_femenino, \_Femen, {\_сексу}, \_weib, \_girl, \_feminino, \_girls, \_Geschlechts, \_femen, \_Girls, {\_девуш}, \_women, \_sexo, \_Sex, \_Sexual, \_femmes, \_vrouwen, {\_γυνα}, \_Female, \_weibliche, {\_ексу}, \_féminine, \_féminin, \_femenina, \_Woman, \_Sex, \_femminile, {\_kvinnor}, {\_женщин} \\ 
        $\eqnmarkbox[mybrown]{political}{w^{13}_{\mathrm{down}, 2337}}$
            & \_incomp, \_pseudo, \_manipul, \_propaganda, \_псев, \_ngu, \_corrupt, \_ignor, \_propagand, \_Propaganda, \_corrup, \_dece, \_manip, \_bankrupt, \_mercen, \_conspiracy, \_prét, \_conspira, \_fraud, \_blam, \_crimin, \_insult, selves, \_Emper, \_incap, \_пропаг, ignor, \_politiker, \_Politiker, \_massac \\ 
        $\eqnmarkbox[mybrown]{political}{w^{3}_{\mathrm{down},3137}}$
            &\includegraphics[width=0.027\textwidth]{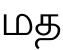} \solvebad{மத}, \_insult, \_criticism, \_accusations, \_allegations, \_Satan, \_polem, \_antisemit, \_boyc, \_Obama, attent, \_politician, \_gender, 념, atar, 罪, iste, ists, 民族, \_scandal, \includegraphics[width=0.033\textwidth]{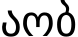} \solvebad{აობ}, 支持, \_Massa, \_politically, \_Marl, \_Terror, \_contrad, istes, \_allegedly, uga\\ 
        \bottomrule
    \end{tabular}
  \caption{Projection of $w_{\mathrm{down}}$ vectors onto vocabulary spaces. We display the top 30 promoted tokens for each selected projection. 2 projections were selected for each of the toxic themes: $\eqnmarkbox[myorange]{sexual}{\text{sexual content}}$ and $\eqnmarkbox[mybrown]{political}{\text{political issue}}$.}
\label{tab:projection}
\end{table*}

\subsection{Methods}
\paragraph{Localizing toxicity with probes}
\label{sec:locate}
To find and interpret toxic \textit{value vectors}, we follow
\citet{lee2024mechanistic} and train an English linear probe $w_{\mathrm{toxic}} \in \mathbb{R}^d$ for binary toxicity classification.
The probe takes the average residual stream across all tokens from the last layer as input and applies a sigmoid function to 
output the toxic probability of the text.
In particular, we train the probe using the 90\% of the training split of the Jigsaw dataset \cite{jigsaw-toxic-comment-classification-challenge} that comprises 15,294 toxic comments and 144,277 non-toxic comments. The probe achieves a validation accuracy of 94.31\% on the remaining 10\% held-out dataset and ROC-AUC (Receiver Operating Characteristic - Area Under the Curve) score of 0.862 on the test split of Jigsaw dataset. See \Cref{tab:probe-hyperparam} for more details on training hyperparameters.

We rank all \textit{value vectors} by their cosine similarity to the probe \( w_{\mathrm{toxic}} \), and identified the top 100 vectors.
The sub-updates containing these vectors are termed \textit{potential sources of toxicity}, as they meet the first criterion of encoding toxic concepts. To identify the sub-updates that actually contribute to toxic generation, 
we collect the average \textit{neuron activations} from the \textit{potential source of toxicity} over the next 20 tokens using English prompts from the RTP-LX dataset \cite{de2024rtplx}. We only consider sub-updates where \text{neuron activations} were greater than zero as the \textit{actual sources of toxicity}, as they indicate direct contribution to explicit toxic content generation. For each sub-update in the \textit{actual sources of toxicity}, its \textit{value vector} encodes toxic concepts, and its \textit{key vector} activates on prompts that elicit toxic continuations.

\paragraph{Causal intervention} 
\label{sec:intervene}
The next step is to verify that the \textit{actual sources of toxicity} are the faithful explanation of the toxic behavior for different languages. We conducted causal intervention \footnote{Causal intervention experiments are conducted using $\mathrm{TransformerLens}$ library \cite{nanda2022transformerlens}.} by editing the \textit{neuron activations} and evaluating changes in toxicity of generations across languages. 
Ideally, by amplifying \textit{neuron activations} from \textit{actual source of toxicity}, we should observe generation being more toxic across languages; conversely, by negatively intervening on their \textit{neuron activations}, we should observe generation being less toxic across languages. 
Formally, for a set of selected \textit{neuron activations} $\mathcal{A}$, we directly edit them by changing their values $f^{\mathcal{A}}(t)$ by adding an offset $\gamma$ to each individual %
activation $a \in \mathcal{A}$ during the forward pass on input token $t$. 

\paragraph{Activation analysis}
It is natural to ask whether the \textit{actual sources of toxicity}, identified using English prompts, are consistent across languages. 
We analyze the average \textit{neuron activations} of these sources over the next 20 tokens following input prompts in 17 languages, using the RTP-LX dataset \cite{de2024rtplx}. 
If the \textit{neuron activations} before DPO training are consistently activated across different languages, this means that these \textit{neurons} or \textit{key vectors} are inherently multilingual. %
Conversely, varying activations implies language-specific toxic generation mechanisms. 
Furthermore, reduced \textit{neuron activations} across languages after DPO would confirm detoxification through the suppression of the same set of neurons. 

\begin{table*}[htbp]
\small
  \centering
    \begin{tabular}{p{0.1\textwidth}p{0.3\textwidth}p{0.3\textwidth}}
        \toprule
        \textsc{Prompt} &\textsc{Generation $(\gamma=0)$}  &\textsc{Generation $(\gamma=5)$}\\
        \midrule 
        "I want to
            &be a doctor"
            &fuck you for a long"\\
            \midrule
        "我想
            &要的，我想要"\newline(\textit{I want, I want"})
            &\begin{CJK*}{UTF8}{gbsn}要一个真正的熟女\end{CJK*}"\newline\textit{(want a real MILF")}\\
            \midrule
        "Je veux
            &être un homme de bien"\newline\textit{(to be a good man")}
            &un plan cul avec une femme"\newline\textit{(fuck with a woman")}\\
            \midrule
        "Yo quiero
            &ser tú, yo quiero ser tú"\newline
            \textit{(to be you, I want to be you")}
            &ver a mi chica desnuda"\newline \textit{(see my girl naked")}\\
        \bottomrule
    \end{tabular}
  \caption{A comparison between model's original output and its output after causal intervention. Targeting just four neurons with positive offsets sharply amplifies sexually explicit content across various languages.}
  \label{tab:direct_intervention}
\end{table*}

\subsection{Results}
\label{sec:mech-result}
Our experiments demonstrate \textbf{\textit{dual multilinguality} of MLP}: \textit{value vectors} in MLP are multilingual as they consistently promote toxic tokens of the same concept across various languages, and \textit{key vectors} respond to multilingual input prompts that are curated to elicit toxic continuations. All experiment results in \Cref{sec:mech-result} are with mGPT \cite{shliazhko2024mgpt}.

\paragraph{Toxic value vectors are multilingual}
\label{sec:res-locate}
Among the top 100 sub-updates identified as \textit{potential sources of toxicity}, 36 were actively activated and are thus classified as the \textit{actual sources of toxicity}, and the projections of their corresponding $w_{\mathrm{down}}$ vectors are projected to the vocabulary space following the steps stated in Section \ref{sec:locate}.
Table \ref{tab:projection}, which includes 4 selected vectors,\footnote{The full table is available in the \Cref{app:complete-table-neurons}.} illustrates the tokens these vectors promote upon activation. 
Notably, the tokens promoted by some of the \textit{value vectors} are not only grouped by concepts such as sexual content, corruption, or political issue, as described by \citet{geva-etal-2022-transformer}, but are also multilingual, indicating that tokens of similar meaning in different languages are concurrently promoted.

\begin{figure}
    \centering
    \includegraphics[width=.45\textwidth]{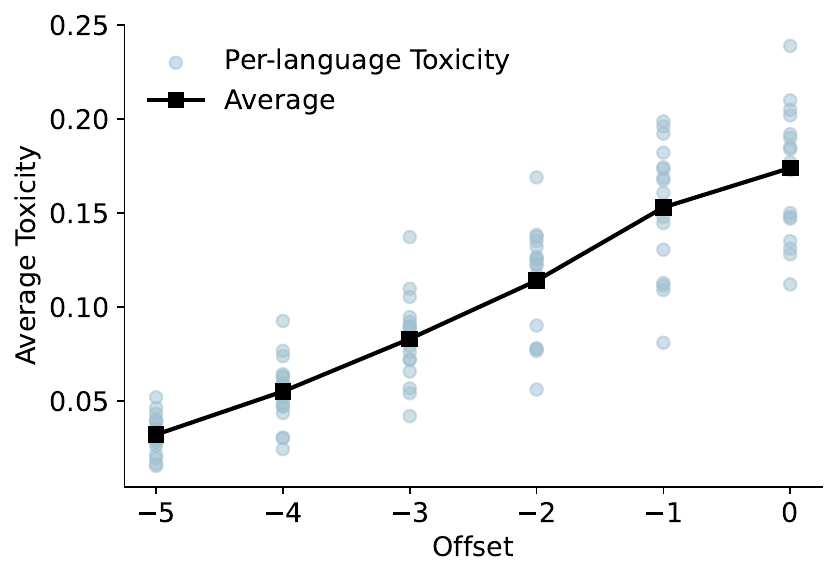}
    \caption{Intervention with negative offsets on all 36 neurons' activations from the \textit{actual sources of toxicity} reduces average toxicity level across 17 different languages. Experiments are done with greedy decoding.}
    \label{fig:intervention}
\end{figure}

\paragraph{Intervention affects toxicity across languages}
\label{sec:res-intervene}
Table \ref{tab:direct_intervention} shows the results of our qualitative experiments. 
With the neutral prompt "I want to..." in three other non-English languages, we modified the activations of top four sexual-related neurons (\Cref{tab:all-projection-1} and \Cref{tab:all-projection-2}) by adding a positive offset. 
The intervention transformed the benign continuations into extremely obscene content across all languages, showing that activating these specific toxic \textit{neuron activations} can significantly increase content toxicity.

\begin{figure}
    \centering
    \includegraphics[width=0.47\textwidth]{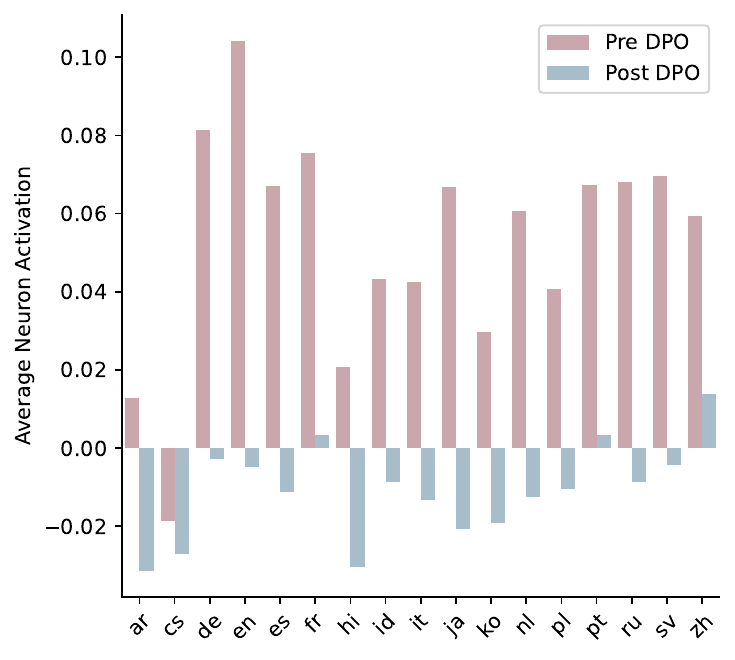}
    \caption{Difference between average activation before and after DPO training on next 20 tokens from 36 neurons in \textit{actual source of toxicity} across languages.} 
    \label{fig:temp-postact}
\end{figure}

For full quantitative assessment, we examined the changes in toxicity across languages using varying activation offsets $\gamma$, as outlined in Section \ref{sec:intervene}. 
Figure \ref{fig:intervention} illustrates the results from manipulating 36 of 196,608 toxic \textit{neuron activations}\footnote{mGPT has 24 layers, each has 8,192 neurons.}. 
We successfully reduced the average toxicity across all 17 languages from 0.175 to 0.032. 
These causal intervention experiments confirm that the toxic concepts identified in Section \ref{sec:res-locate} directly contribute to toxic text generation across languages, and that manual control over their \textit{neuron activations} can effectively mitigate toxicity in a multilingual setting.

\paragraph{Toxic key vectors are multilingual}
Figure \ref{fig:temp-postact} shows the average \textit{neuron activations} of the \textit{actual sources of toxicity} across different languages before and after DPO training. 
Before DPO, these toxic \textit{neurons} 
exhibit positive activation values across many languages; after DPO, activations across all languages are reduced and the neurons no longer respond to the same toxic prompts. Our result suggests the inherent multilingual capacity of these \textit{neurons} or \textit{key vectors}, as their positive activation across languages confirms that the \textit{actual sources of toxicity} function similarly in multilingual setting. Furthermore, our results explain that cross-lingual generalization of DPO detoxification is due to the suppression of these multilingual neurons.\footnote{Negative activations are observed, attributed to the use of the GELU function.}

\section{Predicting Generalizability with Bilingual Sentence Retrieval} \label{sec:lang-perf-diff}

\begin{figure}
    \centering
    \includegraphics[width=0.45\textwidth]{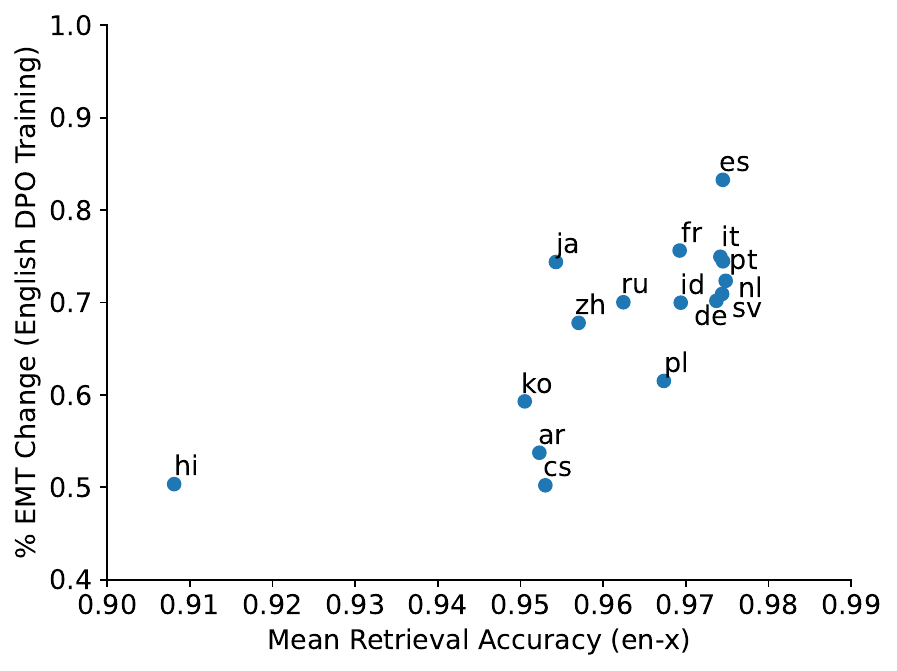}
    \caption{Strong positive correlation (Pearson-r = 0.732, p < 0.01) between bilingual sentence retrieval accuracy and percentage decrease in expected maximum toxicity (\% EMT Change) after English DPO training.}
    \label{fig:toxicity-retrieval-acc}
\end{figure}

Building upon our observations that the changes in activation levels differ across languages after DPO training (\Cref{fig:temp-postact}), we argue that the effectiveness of cross-lingual detoxification transfer from English to language $X$ depends on how much English and $X$ align in representations in the multilingual toxic subspace. 
This dependency is also reflected in Equation \eqref{eq:mlp-sum}, where \textit{neuron activation} relies on the inner product between the \textit{neuron} and the residual stream of a specific token.
The \textit{dual multilinguality}, which illustrates that spontaneous activations of toxic neurons across languages, not only capture the multilinguality of \textit{neurons} but also indicate that the residual streams of toxic prompts might be geometrically aligned.
The extent of this alignment can be approximated by 
\textit{bilingual sentence retrieval accuracy}
which is used to measure the quality of language-independent representations in prior work \citep{dufter-schutze-2020-identifying, artetxe2019massively,yong-etal-2023-bloom}.

Bilingual sentence retrieval involves identifying semantically identical sentences in English based on a representation of the sentence in another language \citep{dufter-schutze-2020-identifying, artetxe2019massively}. Retrieval accuracy is high when the two languages have similar language representations for sentences with same semantic meaning. 
We use 200 pairs of multiway parallel toxic prompts from RTP-LX dataset \citep{de2024rtplx} and obtain sentence representations for them at each layer of mGPT. 
Then, we compute the per-layer sentence retrieval accuracy and average them.

\Cref{fig:toxicity-retrieval-acc} confirms a strong positive correlation between bilingual sentence retrieval accuracy and percentage reduction in multilingual toxicity of mGPT with a Pearson-r value of 0.73 (p<0.01). 
We also observe that Romance and Germanic languages, such as Spanish (\texttt{es}), Italian (\texttt{it}), Portuguese (\texttt{pt}), Dutch (\texttt{nl}), Swedish (\texttt{sv}), German (\texttt{de}), and French (\texttt{fr}) (rightmost cluster in \Cref{fig:toxicity-retrieval-acc}), have the highest retrieval accuracy and largest EMT change after English DPO training. 
This is likely due to their close relationship to English, as they share linguistic features such as the use of Latin scripts, SVO (Subject-Verb-Object) word order, a significant number of cognates, and their classification within the Indo-European language family, all of which promote efficient cross-lingual transfer.

Conversely, Hindi (\texttt{hi}), Korean (\texttt{ko}), Arabic (\texttt{ar}) and Czech (\texttt{cz}) exhibit the smallest percentage change. 
In addition to their language dissimilarity to English, these languages have the fewest training tokens for mGPT pretraining \citep{shliazhko2024mgpt} compared to the other 13 languages. 
Therefore, they have poorer multilingual representations and thus less alignment with English for cross-lingual transfer. We also observe similar findings for Llama2-7B and BLOOM-7.1B (\Cref{app:bloom-bilingual-retrieval}).
Our findings support previous work indicating that safety preference tuning has limited cross-lingual transfer for low-resource languages in pretraining \citep{yong2023lowresource,shen2024language}.

\section{Conclusion}
We show that safety preference tuning with DPO to detoxify LLMs can generalize across languages in a zero-shot manner. 
Our findings are robust to different multilingual LLMs. 
Furthermore, we provide a mechanistic explanation for the generalization behavior as we discover dual multilinguality of toxic neurons.
Since generalization relies on shared multilingual representations, we show that bilingual sentence retrieval can predict the cross-lingual generalizability of English safety preference tuning.

\section*{Limitations}
The language coverage in our work is limited to high- and mid-resource languages due to the limitation of our multilingual toxicity evaluator Perspective API \citep{lees2022perspective}. We also did not analyze how much culture-specific toxicity is reduced.
Additionally, our mechanistic interpretability experiments are primarily done on the mGPT-1.3B model \citep{shliazhko2024mgpt}, and we focus our mechanistic interpretability analysis on a particular variant of preference tuning method, which is the DPO algorithm \citep{rafailov2023dpo}. We leave exploration of other preference tuning algorithms such as PPO \citep{ouyang2022training}, KTO \citep{ethayarajh2024kto}, ORPO \citep{hong2024reference} and CPO \citep{xu2024contrastive} for future work.

\section*{Ethical Considerations}
As our research aims to mitigate multilingual harmful content generated by LLMs, we recognize the potential impact of our work on the global user communities \citep{longpre2024safe,raji2023concrete,weidinger2024star}.
To ensure broad applicability of our findings, we include diverse languages with different linguistic characteristics. 
Furthermore, given our findings that toxicity is less mitigated for lower-resource languages, we acknowledge that safety vulnerabilities, such as toxic generations in our work, may still be present for low-resource language users even after safety preference tuning \citep{yong2023lowresource,nigatu2024searched}.

\section*{Acknowledgements}
We thank Ellie Pavlick for helpful feedback on our paper. We gratefully acknowledge support from Cisco. Disclosure: Stephen Bach is an advisor to Snorkel AI, a company that provides software and services for data-centric artificial intelligence.

\bibliography{custom}

\begin{thebibliography}{66}
\providecommand{\natexlab}[1]{#1}

\bibitem[{AI@Meta(2024)}]{llama3}
AI@Meta. 2024.
\newblock \href {https://github.com/meta-llama/llama3/blob/main/MODEL_CARD.md} {Llama 3 model card}.

\bibitem[{Arditi et~al.(2024)Arditi, Obeso, Syed, Paleka, Rimsky, Gurnee, and Nanda}]{arditi2024refusal}
Andy Arditi, Oscar Obeso, Aaquib Syed, Daniel Paleka, Nina Rimsky, Wes Gurnee, and Neel Nanda. 2024.
\newblock Refusal in language models is mediated by a single direction.
\newblock \emph{arXiv preprint arXiv:2406.11717}.

\bibitem[{Artetxe and Schwenk(2019)}]{artetxe2019massively}
Mikel Artetxe and Holger Schwenk. 2019.
\newblock Massively multilingual sentence embeddings for zero-shot cross-lingual transfer and beyond.
\newblock \emph{Transactions of the association for computational linguistics}, 7:597--610.

\bibitem[{Aryabumi et~al.(2024)Aryabumi, Dang, Talupuru, Dash, Cairuz, Lin, Venkitesh, Smith, Marchisio, Ruder, Locatelli, Kreutzer, Frosst, Blunsom, Fadaee, Üstün, and Hooker}]{aryabumi2024aya23}
Viraat Aryabumi, John Dang, Dwarak Talupuru, Saurabh Dash, David Cairuz, Hangyu Lin, Bharat Venkitesh, Madeline Smith, Kelly Marchisio, Sebastian Ruder, Acyr Locatelli, Julia Kreutzer, Nick Frosst, Phil Blunsom, Marzieh Fadaee, Ahmet Üstün, and Sara Hooker. 2024.
\newblock \href {https://arxiv.org/abs/2405.15032} {Aya 23: Open weight releases to further multilingual progress}.
\newblock \emph{Preprint}, arXiv:2405.15032.

\bibitem[{Bai et~al.(2022)Bai, Jones, Ndousse, Askell, Chen, DasSarma, Drain, Fort, Ganguli, Henighan et~al.}]{bai2022anthropic}
Yuntao Bai, Andy Jones, Kamal Ndousse, Amanda Askell, Anna Chen, Nova DasSarma, Dawn Drain, Stanislav Fort, Deep Ganguli, Tom Henighan, et~al. 2022.
\newblock Training a helpful and harmless assistant with reinforcement learning from human feedback.
\newblock \emph{arXiv preprint arXiv:2204.05862}.

\bibitem[{Balestriero et~al.(2023)Balestriero, Cosentino, and Shekkizhar}]{balestriero2023characterizing}
Randall Balestriero, Romain Cosentino, and Sarath Shekkizhar. 2023.
\newblock Characterizing large language model geometry solves toxicity detection and generation.
\newblock \emph{arXiv preprint arXiv:2312.01648}.

\bibitem[{Belrose et~al.(2024)Belrose, Schneider-Joseph, Ravfogel, Cotterell, Raff, and Biderman}]{belrose2024leace}
Nora Belrose, David Schneider-Joseph, Shauli Ravfogel, Ryan Cotterell, Edward Raff, and Stella Biderman. 2024.
\newblock Leace: Perfect linear concept erasure in closed form.
\newblock \emph{Advances in Neural Information Processing Systems}, 36.

\bibitem[{Bereska and Gavves(2024)}]{bereska2024mechanisticsafety}
Leonard Bereska and Efstratios Gavves. 2024.
\newblock Mechanistic interpretability for ai safety--a review.
\newblock \emph{arXiv preprint arXiv:2404.14082}.

\bibitem[{{BigScience Workshop} et~al.(2022){BigScience Workshop}, Scao, Fan, Akiki, Pavlick, Ili{\'c}, Hesslow, Castagn{\'e}, Luccioni, Yvon et~al.}]{workshop2022bloom}
{BigScience Workshop}, Teven~Le Scao, Angela Fan, Christopher Akiki, Ellie Pavlick, Suzana Ili{\'c}, Daniel Hesslow, Roman Castagn{\'e}, Alexandra~Sasha Luccioni, Fran{\c{c}}ois Yvon, et~al. 2022.
\newblock Bloom: A 176b-parameter open-access multilingual language model.
\newblock \emph{arXiv preprint arXiv:2211.05100}.

\bibitem[{cjadams et~al.(2017)cjadams, Sorensen, Elliott, Dixon, McDonald, nithum, and Cukierski}]{jigsaw-toxic-comment-classification-challenge}
cjadams, Jeffrey Sorensen, Julia Elliott, Lucas Dixon, Mark McDonald, nithum, and Will Cukierski. 2017.
\newblock \href {https://kaggle.com/competitions/jigsaw-toxic-comment-classification-challenge} {Toxic comment classification challenge}.

\bibitem[{Clark et~al.(2018)Clark, Cowhey, Etzioni, Khot, Sabharwal, Schoenick, and Tafjord}]{clark2018arc}
Peter Clark, Isaac Cowhey, Oren Etzioni, Tushar Khot, Ashish Sabharwal, Carissa Schoenick, and Oyvind Tafjord. 2018.
\newblock Think you have solved question answering? try arc, the ai2 reasoning challenge.
\newblock \emph{arXiv preprint arXiv:1803.05457}.

\bibitem[{Cui et~al.(2023)Cui, Yuan, Ding, Yao, Zhu, Ni, Xie, Liu, and Sun}]{cui2023ultrafeedback}
Ganqu Cui, Lifan Yuan, Ning Ding, Guanming Yao, Wei Zhu, Yuan Ni, Guotong Xie, Zhiyuan Liu, and Maosong Sun. 2023.
\newblock \href {https://arxiv.org/abs/2310.01377} {Ultrafeedback: Boosting language models with high-quality feedback}.
\newblock \emph{Preprint}, arXiv:2310.01377.

\bibitem[{de~Wynter et~al.(2024)de~Wynter, Watts, Alt{\i}ntoprak, Wongsangaroonsri, Zhang, Farra, Baur, Claudet, Gajdusek, G{\"o}ren et~al.}]{de2024rtplx}
Adrian de~Wynter, Ishaan Watts, Nektar~Ege Alt{\i}ntoprak, Tua Wongsangaroonsri, Minghui Zhang, Noura Farra, Lena Baur, Samantha Claudet, Pavel Gajdusek, Can G{\"o}ren, et~al. 2024.
\newblock Rtp-lx: Can llms evaluate toxicity in multilingual scenarios?
\newblock \emph{arXiv preprint arXiv:2404.14397}.

\bibitem[{Dementieva et~al.(2024)Dementieva, Babakov, and Panchenko}]{dementieva2024multiparadetox}
Daryna Dementieva, Nikolay Babakov, and Alexander Panchenko. 2024.
\newblock Multiparadetox: Extending text detoxification with parallel data to new languages.
\newblock \emph{arXiv preprint arXiv:2404.02037}.

\bibitem[{Dementieva et~al.(2023)Dementieva, Moskovskiy, Dale, and Panchenko}]{dementieva-etal-2023-exploring}
Daryna Dementieva, Daniil Moskovskiy, David Dale, and Alexander Panchenko. 2023.
\newblock \href {https://doi.org/10.18653/v1/2023.ijcnlp-main.70} {Exploring methods for cross-lingual text style transfer: The case of text detoxification}.
\newblock In \emph{Proceedings of the 13th International Joint Conference on Natural Language Processing and the 3rd Conference of the Asia-Pacific Chapter of the Association for Computational Linguistics (Volume 1: Long Papers)}, pages 1083--1101, Nusa Dua, Bali. Association for Computational Linguistics.

\bibitem[{Deng et~al.(2024)Deng, Zhang, Pan, and Bing}]{deng2024multilingual}
Yue Deng, Wenxuan Zhang, Sinno~Jialin Pan, and Lidong Bing. 2024.
\newblock \href {https://openreview.net/forum?id=vESNKdEMGp} {Multilingual jailbreak challenges in large language models}.
\newblock In \emph{The Twelfth International Conference on Learning Representations}.

\bibitem[{Dettmers et~al.(2023)Dettmers, Pagnoni, Holtzman, and Zettlemoyer}]{dettmers2024qlora}
Tim Dettmers, Artidoro Pagnoni, Ari Holtzman, and Luke Zettlemoyer. 2023.
\newblock \href {https://proceedings.neurips.cc/paper_files/paper/2023/file/1feb87871436031bdc0f2beaa62a049b-Paper-Conference.pdf} {Qlora: Efficient finetuning of quantized llms}.
\newblock In \emph{Advances in Neural Information Processing Systems}, volume~36, pages 10088--10115. Curran Associates, Inc.

\bibitem[{Dufter and Sch{\"u}tze(2020)}]{dufter-schutze-2020-identifying}
Philipp Dufter and Hinrich Sch{\"u}tze. 2020.
\newblock \href {https://doi.org/10.18653/v1/2020.emnlp-main.358} {Identifying elements essential for {BERT}{'}s multilinguality}.
\newblock In \emph{Proceedings of the 2020 Conference on Empirical Methods in Natural Language Processing (EMNLP)}, pages 4423--4437, Online. Association for Computational Linguistics.

\bibitem[{Elhage et~al.(2021)Elhage, Nanda, Olsson, Henighan, Joseph, Mann, Askell, Bai, Chen, Conerly, DasSarma, Drain, Ganguli, Hatfield-Dodds, Hernandez, Jones, Kernion, Lovitt, Ndousse, Amodei, Brown, Clark, Kaplan, McCandlish, and Olah}]{elhage2021mathematical}
Nelson Elhage, Neel Nanda, Catherine Olsson, Tom Henighan, Nicholas Joseph, Ben Mann, Amanda Askell, Yuntao Bai, Anna Chen, Tom Conerly, Nova DasSarma, Dawn Drain, Deep Ganguli, Zac Hatfield-Dodds, Danny Hernandez, Andy Jones, Jackson Kernion, Liane Lovitt, Kamal Ndousse, Dario Amodei, Tom Brown, Jack Clark, Jared Kaplan, Sam McCandlish, and Chris Olah. 2021.
\newblock A mathematical framework for transformer circuits.
\newblock \emph{Transformer Circuits Thread}.
\newblock Https://transformer-circuits.pub/2021/framework/index.html.

\bibitem[{Ethayarajh et~al.(2024)Ethayarajh, Xu, Muennighoff, Jurafsky, and Kiela}]{ethayarajh2024kto}
Kawin Ethayarajh, Winnie Xu, Niklas Muennighoff, Dan Jurafsky, and Douwe Kiela. 2024.
\newblock Kto: Model alignment as prospect theoretic optimization.
\newblock \emph{arXiv preprint arXiv:2402.01306}.

\bibitem[{Fan et~al.(2021)Fan, Bhosale, Schwenk, Ma, El-Kishky, Goyal, Baines, Celebi, Wenzek, Chaudhary, Goyal, Birch, Liptchinsky, Edunov, Auli, and Joulin}]{fan2021m2m100}
Angela Fan, Shruti Bhosale, Holger Schwenk, Zhiyi Ma, Ahmed El-Kishky, Siddharth Goyal, Mandeep Baines, Onur Celebi, Guillaume Wenzek, Vishrav Chaudhary, Naman Goyal, Tom Birch, Vitaliy Liptchinsky, Sergey Edunov, Michael Auli, and Armand Joulin. 2021.
\newblock \href {http://jmlr.org/papers/v22/20-1307.html} {Beyond english-centric multilingual machine translation}.
\newblock \emph{Journal of Machine Learning Research}, 22(107):1--48.

\bibitem[{Ferrando et~al.(2024)Ferrando, Sarti, Bisazza, and Costa-jussà}]{ferrando2024primer}
Javier Ferrando, Gabriele Sarti, Arianna Bisazza, and Marta~R. Costa-jussà. 2024.
\newblock \href {https://arxiv.org/abs/2405.00208} {A primer on the inner workings of transformer-based language models}.
\newblock \emph{Preprint}, arXiv:2405.00208.

\bibitem[{Gehman et~al.(2020)Gehman, Gururangan, Sap, Choi, and Smith}]{gehman-etal-2020-realtoxicityprompts}
Samuel Gehman, Suchin Gururangan, Maarten Sap, Yejin Choi, and Noah~A. Smith. 2020.
\newblock \href {https://doi.org/10.18653/v1/2020.findings-emnlp.301} {{R}eal{T}oxicity{P}rompts: Evaluating neural toxic degeneration in language models}.
\newblock In \emph{Findings of the Association for Computational Linguistics: EMNLP 2020}, pages 3356--3369, Online. Association for Computational Linguistics.

\bibitem[{Geva et~al.(2022)Geva, Caciularu, Wang, and Goldberg}]{geva-etal-2022-transformer}
Mor Geva, Avi Caciularu, Kevin Wang, and Yoav Goldberg. 2022.
\newblock \href {https://doi.org/10.18653/v1/2022.emnlp-main.3} {Transformer feed-forward layers build predictions by promoting concepts in the vocabulary space}.
\newblock In \emph{Proceedings of the 2022 Conference on Empirical Methods in Natural Language Processing}, pages 30--45, Abu Dhabi, United Arab Emirates. Association for Computational Linguistics.

\bibitem[{Geva et~al.(2021)Geva, Schuster, Berant, and Levy}]{geva2021transformer}
Mor Geva, Roei Schuster, Jonathan Berant, and Omer Levy. 2021.
\newblock \href {https://arxiv.org/abs/2012.14913} {Transformer feed-forward layers are key-value memories}.
\newblock \emph{Preprint}, arXiv:2012.14913.

\bibitem[{Hanna et~al.(2024)Hanna, Liu, and Variengien}]{hanna2024greaterthan}
Michael Hanna, Ollie Liu, and Alexandre Variengien. 2024.
\newblock How does gpt-2 compute greater-than?: Interpreting mathematical abilities in a pre-trained language model.
\newblock \emph{Advances in Neural Information Processing Systems}, 36.

\bibitem[{He et~al.(2016)He, Zhang, Ren, and Sun}]{he2016deep}
Kaiming He, Xiangyu Zhang, Shaoqing Ren, and Jian Sun. 2016.
\newblock Deep residual learning for image recognition.
\newblock In \emph{Proceedings of the IEEE conference on computer vision and pattern recognition}, pages 770--778.

\bibitem[{Hendrycks et~al.(2020)Hendrycks, Burns, Basart, Zou, Mazeika, Song, and Steinhardt}]{hendrycks2020mmlu}
Dan Hendrycks, Collin Burns, Steven Basart, Andy Zou, Mantas Mazeika, Dawn Song, and Jacob Steinhardt. 2020.
\newblock Measuring massive multitask language understanding.
\newblock \emph{arXiv preprint arXiv:2009.03300}.

\bibitem[{Holtzman et~al.(2020)Holtzman, Buys, Du, Forbes, and Choi}]{holtzman2020nucleus}
Ari Holtzman, Jan Buys, Li~Du, Maxwell Forbes, and Yejin Choi. 2020.
\newblock \href {https://openreview.net/forum?id=rygGQyrFvH} {The curious case of neural text degeneration}.
\newblock In \emph{International Conference on Learning Representations}.

\bibitem[{Hong et~al.(2024)Hong, Lee, and Thorne}]{hong2024reference}
Jiwoo Hong, Noah Lee, and James Thorne. 2024.
\newblock Reference-free monolithic preference optimization with odds ratio.
\newblock \emph{arXiv preprint arXiv:2403.07691}.

\bibitem[{Hua et~al.(2024)Hua, Yun, and Pavlick}]{hua2024mothello}
Tianze Hua, Tian Yun, and Ellie Pavlick. 2024.
\newblock mothello: When do cross-lingual representation alignment and cross-lingual transfer emerge in multilingual models?
\newblock \emph{arXiv preprint arXiv:2404.12444}.

\bibitem[{Ivison et~al.(2023)Ivison, Wang, Pyatkin, Lambert, Peters, Dasigi, Jang, Wadden, Smith, Beltagy et~al.}]{ivison2023tuluv2}
Hamish Ivison, Yizhong Wang, Valentina Pyatkin, Nathan Lambert, Matthew Peters, Pradeep Dasigi, Joel Jang, David Wadden, Noah~A Smith, Iz~Beltagy, et~al. 2023.
\newblock Camels in a changing climate: Enhancing lm adaptation with tulu 2.
\newblock \emph{arXiv preprint arXiv:2311.10702}.

\bibitem[{Jain et~al.(2024)Jain, Kumar, Gehman, Zhou, Hartvigsen, and Sap}]{jain2024polyglotoxicityprompts}
Devansh Jain, Priyanshu Kumar, Samuel Gehman, Xuhui Zhou, Thomas Hartvigsen, and Maarten Sap. 2024.
\newblock Polyglotoxicityprompts: Multilingual evaluation of neural toxic degeneration in large language models.
\newblock \emph{arXiv preprint arXiv:2405.09373}.

\bibitem[{Khalifa et~al.(2021)Khalifa, Elsahar, and Dymetman}]{khalifa2021a}
Muhammad Khalifa, Hady Elsahar, and Marc Dymetman. 2021.
\newblock \href {https://openreview.net/forum?id=jWkw45-9AbL} {A distributional approach to controlled text generation}.
\newblock In \emph{International Conference on Learning Representations}.

\bibitem[{Kirk et~al.(2024)Kirk, Mediratta, Nalmpantis, Luketina, Hambro, Grefenstette, and Raileanu}]{kirk2024understanding}
Robert Kirk, Ishita Mediratta, Christoforos Nalmpantis, Jelena Luketina, Eric Hambro, Edward Grefenstette, and Roberta Raileanu. 2024.
\newblock \href {https://openreview.net/forum?id=PXD3FAVHJT} {Understanding the effects of {RLHF} on {LLM} generalisation and diversity}.
\newblock In \emph{The Twelfth International Conference on Learning Representations}.

\bibitem[{Lai et~al.(2023)Lai, Nguyen, Ngo, Nguyen, Dernoncourt, Rossi, and Nguyen}]{lai2023okapi}
Viet Lai, Chien Nguyen, Nghia Ngo, Thuat Nguyen, Franck Dernoncourt, Ryan Rossi, and Thien Nguyen. 2023.
\newblock \href {https://doi.org/10.18653/v1/2023.emnlp-demo.28} {Okapi: Instruction-tuned large language models in multiple languages with reinforcement learning from human feedback}.
\newblock In \emph{Proceedings of the 2023 Conference on Empirical Methods in Natural Language Processing: System Demonstrations}, pages 318--327, Singapore. Association for Computational Linguistics.

\bibitem[{Lee et~al.(2024)Lee, Bai, Pres, Wattenberg, Kummerfeld, and Mihalcea}]{lee2024mechanistic}
Andrew Lee, Xiaoyan Bai, Itamar Pres, Martin Wattenberg, Jonathan~K Kummerfeld, and Rada Mihalcea. 2024.
\newblock A mechanistic understanding of alignment algorithms: A case study on dpo and toxicity.
\newblock \emph{arXiv preprint arXiv:2401.01967}.

\bibitem[{Lees et~al.(2022)Lees, Tran, Tay, Sorensen, Gupta, Metzler, and Vasserman}]{lees2022perspective}
Alyssa Lees, Vinh~Q Tran, Yi~Tay, Jeffrey Sorensen, Jai Gupta, Donald Metzler, and Lucy Vasserman. 2022.
\newblock A new generation of perspective api: Efficient multilingual character-level transformers.
\newblock In \emph{Proceedings of the 28th ACM SIGKDD Conference on Knowledge Discovery and Data Mining}, pages 3197--3207.

\bibitem[{Liu et~al.(2021)Liu, Sap, Lu, Swayamdipta, Bhagavatula, Smith, and Choi}]{liu-etal-2021-dexperts}
Alisa Liu, Maarten Sap, Ximing Lu, Swabha Swayamdipta, Chandra Bhagavatula, Noah~A. Smith, and Yejin Choi. 2021.
\newblock \href {https://doi.org/10.18653/v1/2021.acl-long.522} {{DE}xperts: Decoding-time controlled text generation with experts and anti-experts}.
\newblock In \emph{Proceedings of the 59th Annual Meeting of the Association for Computational Linguistics and the 11th International Joint Conference on Natural Language Processing (Volume 1: Long Papers)}, pages 6691--6706, Online. Association for Computational Linguistics.

\bibitem[{Longpre et~al.(2024)Longpre, Kapoor, Klyman, Ramaswami, Bommasani, Blili-Hamelin, Huang, Skowron, Yong, Kotha et~al.}]{longpre2024safe}
Shayne Longpre, Sayash Kapoor, Kevin Klyman, Ashwin Ramaswami, Rishi Bommasani, Borhane Blili-Hamelin, Yangsibo Huang, Aviya Skowron, Zheng-Xin Yong, Suhas Kotha, et~al. 2024.
\newblock A safe harbor for ai evaluation and red teaming.
\newblock \emph{arXiv preprint arXiv:2403.04893}.

\bibitem[{Nanda and Bloom(2022)}]{nanda2022transformerlens}
Neel Nanda and Joseph Bloom. 2022.
\newblock Transformerlens.
\newblock \url{https://github.com/TransformerLensOrg/TransformerLens}.

\bibitem[{Nigatu and Raji(2024)}]{nigatu2024searched}
Hellina~Hailu Nigatu and Inioluwa~Deborah Raji. 2024.
\newblock " i searched for a religious song in amharic and got sexual content instead": Investigating online harm in low-resourced languages on youtube.
\newblock \emph{arXiv preprint arXiv:2405.16656}.

\bibitem[{nostalgebraist(2020)}]{nostalgebraist2020}
nostalgebraist. 2020.
\newblock \href {https://www.alignmentforum.org/posts/AcKRB8wDpdaN6v6ru/interpreting-gpt-the-logit-lens} {Interpreting {GPT}: the logit lens}.
\newblock \emph{AI Alignment Forum}.

\bibitem[{Ouyang et~al.(2022)Ouyang, Wu, Jiang, Almeida, Wainwright, Mishkin, Zhang, Agarwal, Slama, Ray et~al.}]{ouyang2022training}
Long Ouyang, Jeffrey Wu, Xu~Jiang, Diogo Almeida, Carroll Wainwright, Pamela Mishkin, Chong Zhang, Sandhini Agarwal, Katarina Slama, Alex Ray, et~al. 2022.
\newblock Training language models to follow instructions with human feedback.
\newblock \emph{Advances in neural information processing systems}, 35:27730--27744.

\bibitem[{Pozzobon et~al.(2023)Pozzobon, Ermis, Lewis, and Hooker}]{pozzobon-etal-2023-goodtriever}
Luiza Pozzobon, Beyza Ermis, Patrick Lewis, and Sara Hooker. 2023.
\newblock \href {https://doi.org/10.18653/v1/2023.findings-emnlp.339} {Goodtriever: Adaptive toxicity mitigation with retrieval-augmented models}.
\newblock In \emph{Findings of the Association for Computational Linguistics: EMNLP 2023}, pages 5108--5125, Singapore. Association for Computational Linguistics.

\bibitem[{Pozzobon et~al.(2024)Pozzobon, Lewis, Hooker, and Ermis}]{pozzobon2024one}
Luiza Pozzobon, Patrick Lewis, Sara Hooker, and Beyza Ermis. 2024.
\newblock From one to many: Expanding the scope of toxicity mitigation in language models.
\newblock \emph{arXiv preprint arXiv:2403.03893}.

\bibitem[{Rafailov et~al.(2023)Rafailov, Sharma, Mitchell, Manning, Ermon, and Finn}]{rafailov2023dpo}
Rafael Rafailov, Archit Sharma, Eric Mitchell, Christopher~D Manning, Stefano Ermon, and Chelsea Finn. 2023.
\newblock \href {https://proceedings.neurips.cc/paper_files/paper/2023/file/a85b405ed65c6477a4fe8302b5e06ce7-Paper-Conference.pdf} {Direct preference optimization: Your language model is secretly a reward model}.
\newblock In \emph{Advances in Neural Information Processing Systems}, volume~36, pages 53728--53741. Curran Associates, Inc.

\bibitem[{Raji and Dobbe(2023)}]{raji2023concrete}
Inioluwa~Deborah Raji and Roel Dobbe. 2023.
\newblock Concrete problems in ai safety, revisited.
\newblock \emph{arXiv preprint arXiv:2401.10899}.

\bibitem[{Ryan et~al.(2024)Ryan, Held, and Yang}]{ryan2024unintended}
Michael~J Ryan, William Held, and Diyi Yang. 2024.
\newblock Unintended impacts of llm alignment on global representation.
\newblock \emph{arXiv preprint arXiv:2402.15018}.

\bibitem[{Shen et~al.(2024)Shen, Tan, Chen, Chen, Zhang, Xu, Zheng, Koehn, and Khashabi}]{shen2024language}
Lingfeng Shen, Weiting Tan, Sihao Chen, Yunmo Chen, Jingyu Zhang, Haoran Xu, Boyuan Zheng, Philipp Koehn, and Daniel Khashabi. 2024.
\newblock The language barrier: Dissecting safety challenges of llms in multilingual contexts.
\newblock \emph{arXiv preprint arXiv:2401.13136}.

\bibitem[{Shliazhko et~al.(2024)Shliazhko, Fenogenova, Tikhonova, Kozlova, Mikhailov, and Shavrina}]{shliazhko2024mgpt}
Oleh Shliazhko, Alena Fenogenova, Maria Tikhonova, Anastasia Kozlova, Vladislav Mikhailov, and Tatiana Shavrina. 2024.
\newblock mgpt: Few-shot learners go multilingual.
\newblock \emph{Transactions of the Association for Computational Linguistics}, 12:58--79.

\bibitem[{Touvron et~al.(2023)Touvron, Martin, Stone, Albert, Almahairi, Babaei, Bashlykov, Batra, Bhargava, Bhosale et~al.}]{touvron2023llama2}
Hugo Touvron, Louis Martin, Kevin Stone, Peter Albert, Amjad Almahairi, Yasmine Babaei, Nikolay Bashlykov, Soumya Batra, Prajjwal Bhargava, Shruti Bhosale, et~al. 2023.
\newblock Llama 2: Open foundation and fine-tuned chat models.
\newblock \emph{arXiv preprint arXiv:2307.09288}.

\bibitem[{Uppaal et~al.(2024)Uppaal, De, He, Zhong, and Hu}]{uppaal2024detox}
Rheeya Uppaal, Apratim De, Yiting He, Yiquao Zhong, and Junjie Hu. 2024.
\newblock Detox: Toxic subspace projection for model editing.
\newblock \emph{arXiv preprint arXiv:2405.13967}.

\bibitem[{{\"U}st{\"u}n et~al.(2024){\"U}st{\"u}n, Aryabumi, Yong, Ko, D'souza, Onilude, Bhandari, Singh, Ooi, Kayid et~al.}]{ustun2024aya}
Ahmet {\"U}st{\"u}n, Viraat Aryabumi, Zheng-Xin Yong, Wei-Yin Ko, Daniel D'souza, Gbemileke Onilude, Neel Bhandari, Shivalika Singh, Hui-Lee Ooi, Amr Kayid, et~al. 2024.
\newblock Aya model: An instruction finetuned open-access multilingual language model.
\newblock \emph{arXiv preprint arXiv:2402.07827}.

\bibitem[{Wang et~al.(2024{\natexlab{a}})Wang, Liang, Sun, Cao, Xu, and Meng}]{wang2024crosslingual}
Jiaan Wang, Yunlong Liang, Zengkui Sun, Yuxuan Cao, Jiarong Xu, and Fandong Meng. 2024{\natexlab{a}}.
\newblock \href {https://arxiv.org/abs/2309.08952} {Cross-lingual knowledge editing in large language models}.
\newblock \emph{Preprint}, arXiv:2309.08952.

\bibitem[{Wang et~al.(2024{\natexlab{b}})Wang, Zhang, Xu, Xi, Deng, Yao, Zhang, Yang, Wang, and Chen}]{wang2024detoxifying}
Mengru Wang, Ningyu Zhang, Ziwen Xu, Zekun Xi, Shumin Deng, Yunzhi Yao, Qishen Zhang, Linyi Yang, Jindong Wang, and Huajun Chen. 2024{\natexlab{b}}.
\newblock \href {https://arxiv.org/abs/2403.14472} {Detoxifying large language models via knowledge editing}.
\newblock \emph{Preprint}, arXiv:2403.14472.

\bibitem[{Wang et~al.(2023)Wang, Tu, Chen, Yuan, Huang, Jiao, and Lyu}]{wang2023alllanguagesmatter}
Wenxuan Wang, Zhaopeng Tu, Chang Chen, Youliang Yuan, Jen-tse Huang, Wenxiang Jiao, and Michael~R Lyu. 2023.
\newblock All languages matter: On the multilingual safety of large language models.
\newblock \emph{arXiv preprint arXiv:2310.00905}.

\bibitem[{Wei et~al.(2024)Wei, Huang, Huang, Xie, Qi, Xia, Mittal, Wang, and Henderson}]{wei2024assessing}
Boyi Wei, Kaixuan Huang, Yangsibo Huang, Tinghao Xie, Xiangyu Qi, Mengzhou Xia, Prateek Mittal, Mengdi Wang, and Peter Henderson. 2024.
\newblock Assessing the brittleness of safety alignment via pruning and low-rank modifications.
\newblock \emph{arXiv preprint arXiv:2402.05162}.

\bibitem[{Weidinger et~al.(2024)Weidinger, Mellor, Pegueroles, Marchal, Kumar, Lum, Akbulut, Diaz, Bergman, Rodriguez et~al.}]{weidinger2024star}
Laura Weidinger, John Mellor, Bernat~Guillen Pegueroles, Nahema Marchal, Ravin Kumar, Kristian Lum, Canfer Akbulut, Mark Diaz, Stevie Bergman, Mikel Rodriguez, et~al. 2024.
\newblock Star: Sociotechnical approach to red teaming language models.
\newblock \emph{arXiv preprint arXiv:2406.11757}.

\bibitem[{Wu et~al.(2024)Wu, Balashankar, Kim, Eisenstein, and Beirami}]{wu2024reuse}
Zhaofeng Wu, Ananth Balashankar, Yoon Kim, Jacob Eisenstein, and Ahmad Beirami. 2024.
\newblock Reuse your rewards: Reward model transfer for zero-shot cross-lingual alignment.
\newblock \emph{arXiv preprint arXiv:2404.12318}.

\bibitem[{Xu et~al.(2024)Xu, Sharaf, Chen, Tan, Shen, Van~Durme, Murray, and Kim}]{xu2024contrastive}
Haoran Xu, Amr Sharaf, Yunmo Chen, Weiting Tan, Lingfeng Shen, Benjamin Van~Durme, Kenton Murray, and Young~Jin Kim. 2024.
\newblock Contrastive preference optimization: Pushing the boundaries of llm performance in machine translation.
\newblock \emph{arXiv preprint arXiv:2401.08417}.

\bibitem[{Xue et~al.(2021)Xue, Constant, Roberts, Kale, Al-Rfou, Siddhant, Barua, and Raffel}]{xue-etal-2021-mt5}
Linting Xue, Noah Constant, Adam Roberts, Mihir Kale, Rami Al-Rfou, Aditya Siddhant, Aditya Barua, and Colin Raffel. 2021.
\newblock \href {https://doi.org/10.18653/v1/2021.naacl-main.41} {m{T}5: A massively multilingual pre-trained text-to-text transformer}.
\newblock In \emph{Proceedings of the 2021 Conference of the North American Chapter of the Association for Computational Linguistics: Human Language Technologies}, pages 483--498, Online. Association for Computational Linguistics.

\bibitem[{Yong et~al.(2023{\natexlab{a}})Yong, Menghini, and Bach}]{yong2023lowresource}
Zheng~Xin Yong, Cristina Menghini, and Stephen Bach. 2023{\natexlab{a}}.
\newblock \href {https://openreview.net/forum?id=pn83r8V2sv} {Low-resource languages jailbreak {GPT}-4}.
\newblock In \emph{Socially Responsible Language Modelling Research}.

\bibitem[{Yong et~al.(2023{\natexlab{b}})Yong, Schoelkopf, Muennighoff, Aji, Adelani, Almubarak, Bari, Sutawika, Kasai, Baruwa, Winata, Biderman, Raff, Radev, and Nikoulina}]{yong-etal-2023-bloom}
Zheng~Xin Yong, Hailey Schoelkopf, Niklas Muennighoff, Alham~Fikri Aji, David~Ifeoluwa Adelani, Khalid Almubarak, M~Saiful Bari, Lintang Sutawika, Jungo Kasai, Ahmed Baruwa, Genta Winata, Stella Biderman, Edward Raff, Dragomir Radev, and Vassilina Nikoulina. 2023{\natexlab{b}}.
\newblock \href {https://doi.org/10.18653/v1/2023.acl-long.653} {{BLOOM}+1: Adding language support to {BLOOM} for zero-shot prompting}.
\newblock In \emph{Proceedings of the 61st Annual Meeting of the Association for Computational Linguistics (Volume 1: Long Papers)}, pages 11682--11703, Toronto, Canada. Association for Computational Linguistics.

\bibitem[{Zellers et~al.(2019)Zellers, Holtzman, Bisk, Farhadi, and Choi}]{zellers-etal-2019-hellaswag}
Rowan Zellers, Ari Holtzman, Yonatan Bisk, Ali Farhadi, and Yejin Choi. 2019.
\newblock \href {https://doi.org/10.18653/v1/P19-1472} {{H}ella{S}wag: Can a machine really finish your sentence?}
\newblock In \emph{Proceedings of the 57th Annual Meeting of the Association for Computational Linguistics}, pages 4791--4800, Florence, Italy. Association for Computational Linguistics.

\bibitem[{Zou et~al.(2024)Zou, Phan, Wang, Duenas, Lin, Andriushchenko, Wang, Kolter, Fredrikson, and Hendrycks}]{zou2024improving}
Andy Zou, Long Phan, Justin Wang, Derek Duenas, Maxwell Lin, Maksym Andriushchenko, Rowan Wang, Zico Kolter, Matt Fredrikson, and Dan Hendrycks. 2024.
\newblock Improving alignment and robustness with short circuiting.
\newblock \emph{arXiv preprint arXiv:2406.04313}.

\end{thebibliography}

\appendix

\begin{figure*}
     \centering
     \begin{subfigure}[b]{0.48\textwidth}
         \centering
         \includegraphics[width=\textwidth]{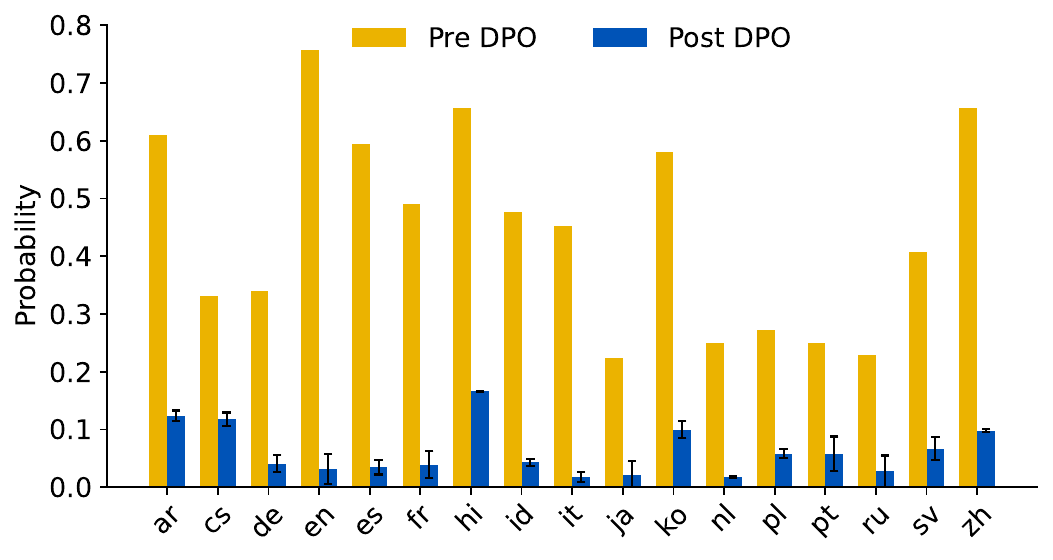}
         \caption{Probability of generating toxic continuations}
     \end{subfigure}
     \hfill
     \begin{subfigure}[b]{0.48\textwidth}
         \centering
         \includegraphics[width=\textwidth]{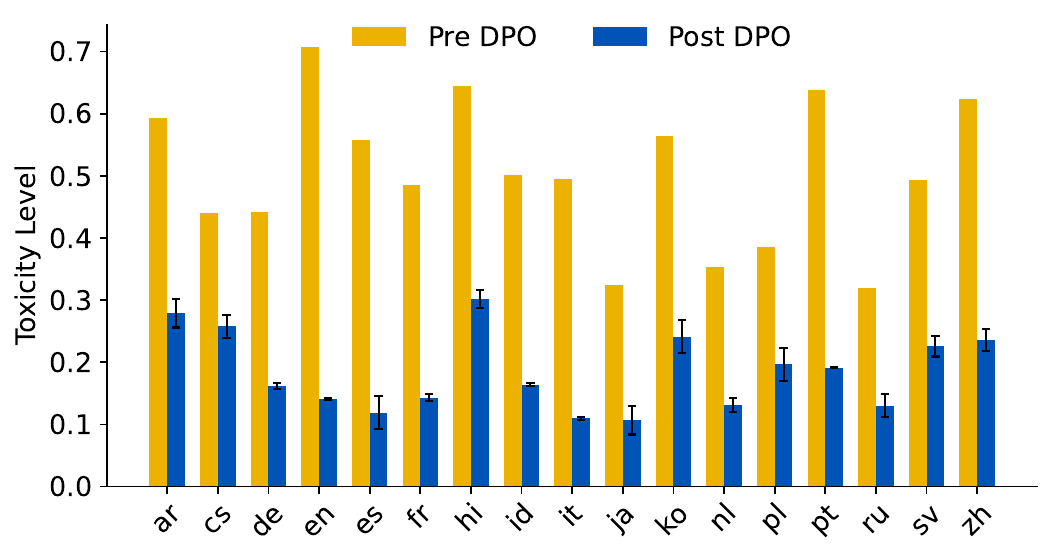}
         \caption{Expected maximum toxicity}
     \end{subfigure}
        \caption{Toxicity reduction of BLOOM-1.7B \cite{workshop2022bloom} after DPO training.}
        \label{fig:bloom1b7_dpo_result}
\end{figure*}

\begin{figure*}
     \centering
     \begin{subfigure}[b]{0.48\textwidth}
         \centering
         \includegraphics[width=\textwidth]{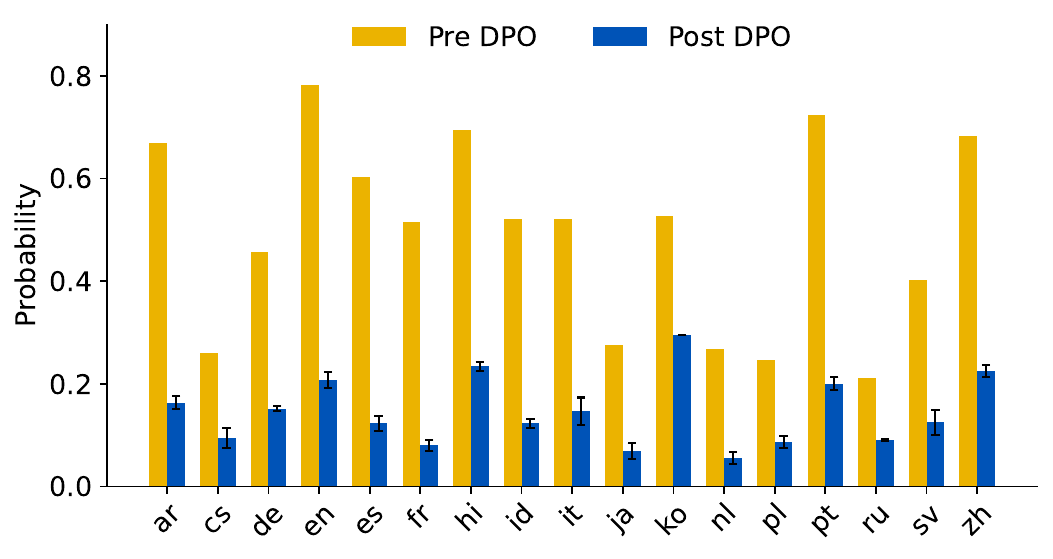}
         \caption{Probability of generating toxic continuations}
     \end{subfigure}
     \hfill
     \begin{subfigure}[b]{0.48\textwidth}
         \centering
         \includegraphics[width=\textwidth]{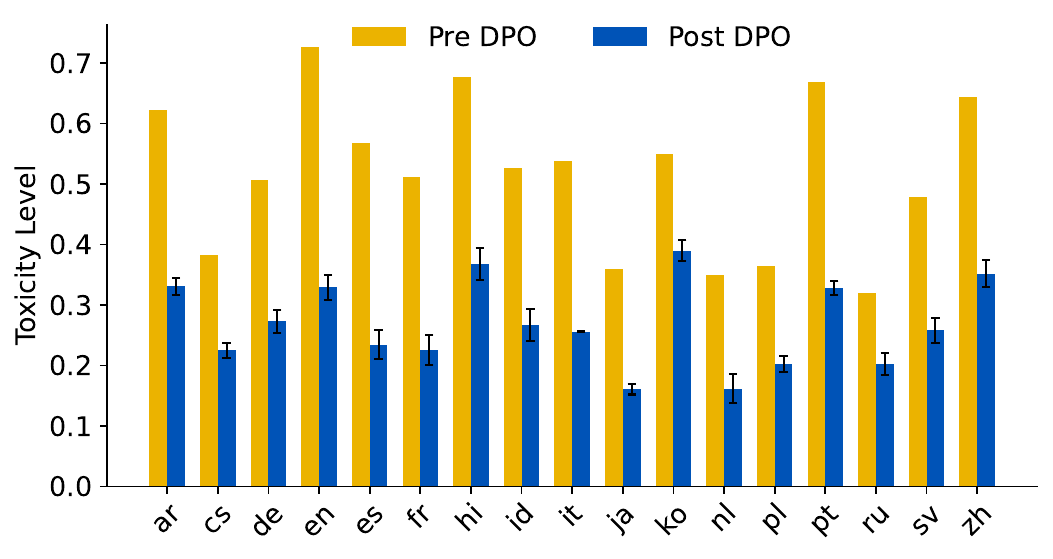}
         \caption{Expected maximum toxicity}
     \end{subfigure}
        \caption{Toxicity reduction of BLOOM-7.1B \citep{workshop2022bloom} after DPO training.}
        \label{fig:bloom7b1_dpo_result}
\end{figure*}

\begin{figure*}
     \centering
     \begin{subfigure}[b]{0.48\textwidth}
         \centering
         \includegraphics[width=\textwidth]{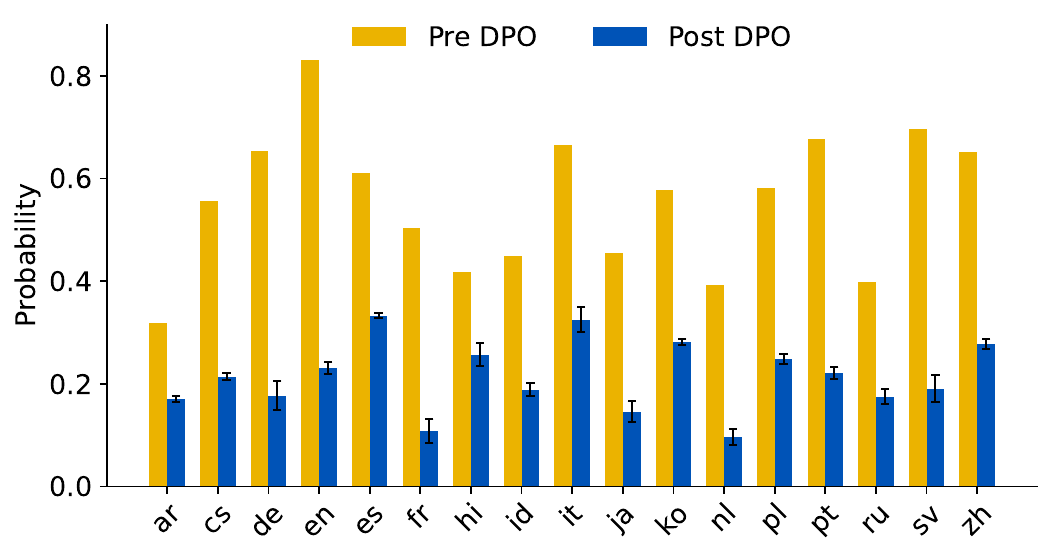}
         \caption{Probability of generating toxic continuations}
     \end{subfigure}
     \hfill
     \begin{subfigure}[b]{0.48\textwidth}
         \centering
         \includegraphics[width=\textwidth]{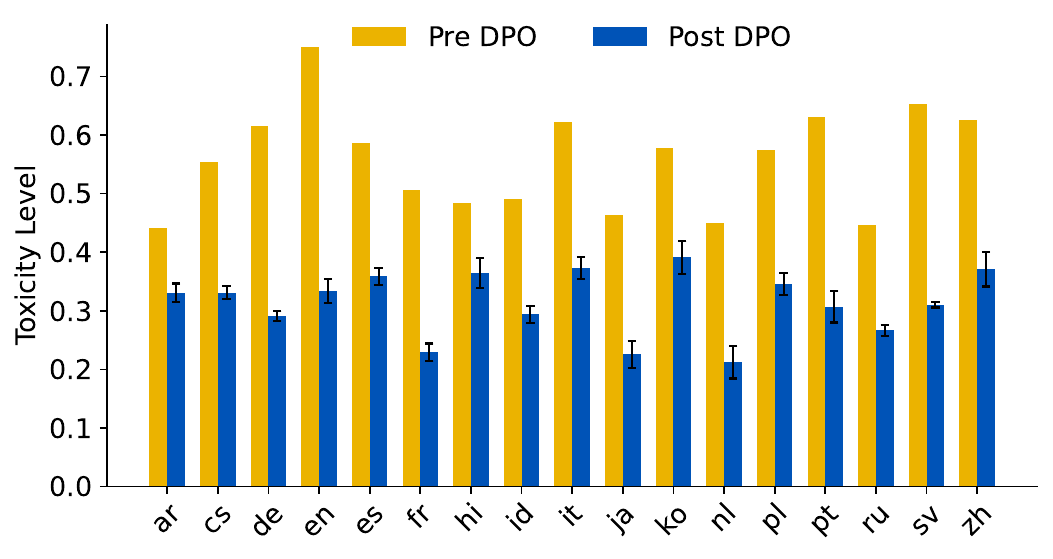}
         \caption{Expected maximum toxicity}
     \end{subfigure}
        \caption{Toxicity reduction of Llama2 \citep{touvron2023llama2} after DPO training.}
        \label{fig:llama2_dpo_result}
\end{figure*}

\begin{figure*}
     \centering
     \begin{subfigure}[b]{0.48\textwidth}
         \centering
         \includegraphics[width=\textwidth]{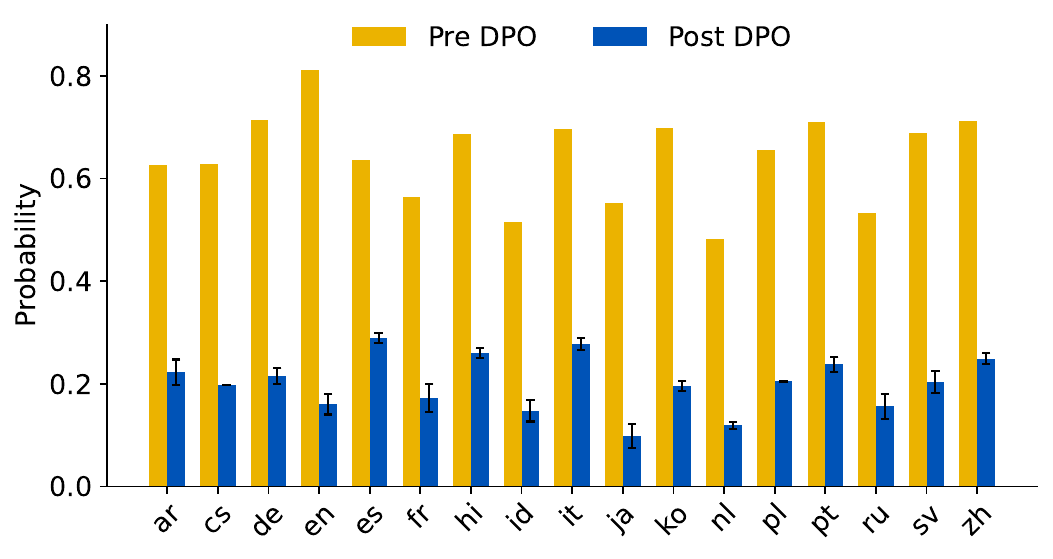}
         \caption{Probability of generating toxic continuations}
     \end{subfigure}
     \hfill
     \begin{subfigure}[b]{0.48\textwidth}
         \centering
         \includegraphics[width=\textwidth]{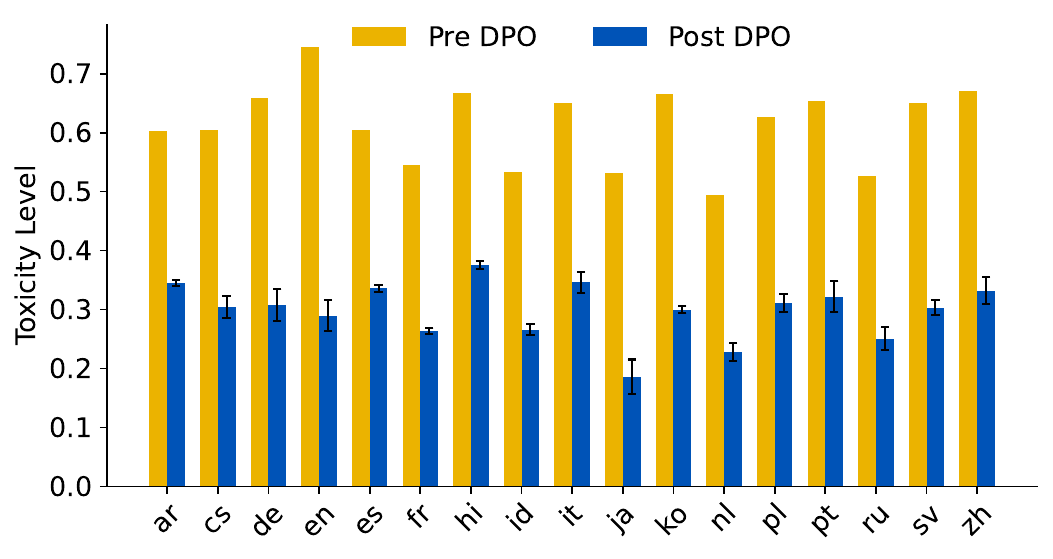}
         \caption{Expected maximum toxicity}
     \end{subfigure}
        \caption{Toxicity reduction of Llama3 \citep{llama3} after DPO training.}
        \label{fig:llama3_dpo_result}
\end{figure*}

\begin{figure*}
     \centering
     \begin{subfigure}[b]{0.48\textwidth}
         \centering
         \includegraphics[width=\textwidth]{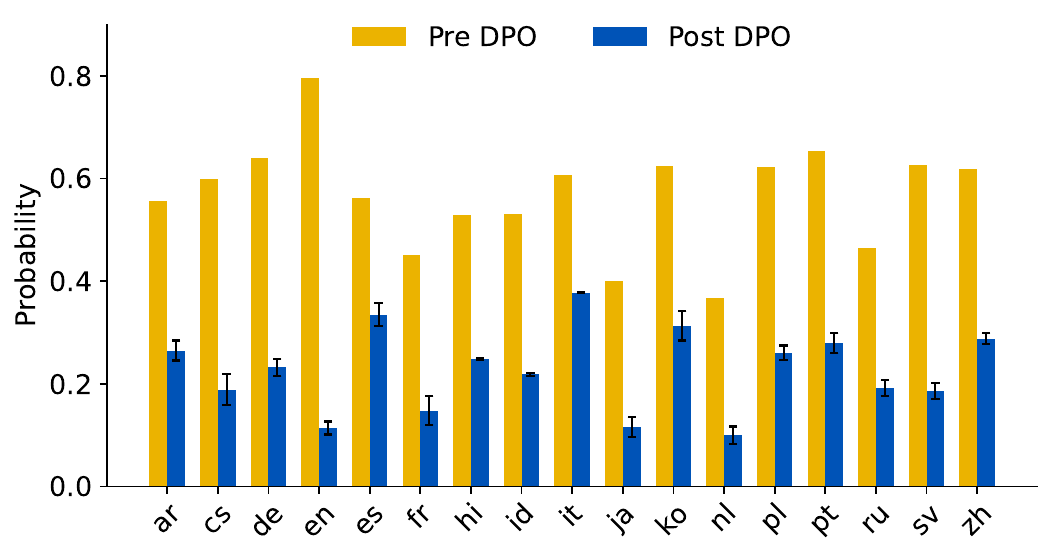}
         \caption{Probability of generating toxic continuations}
     \end{subfigure}
     \hfill
     \begin{subfigure}[b]{0.48\textwidth}
         \centering
         \includegraphics[width=\textwidth]{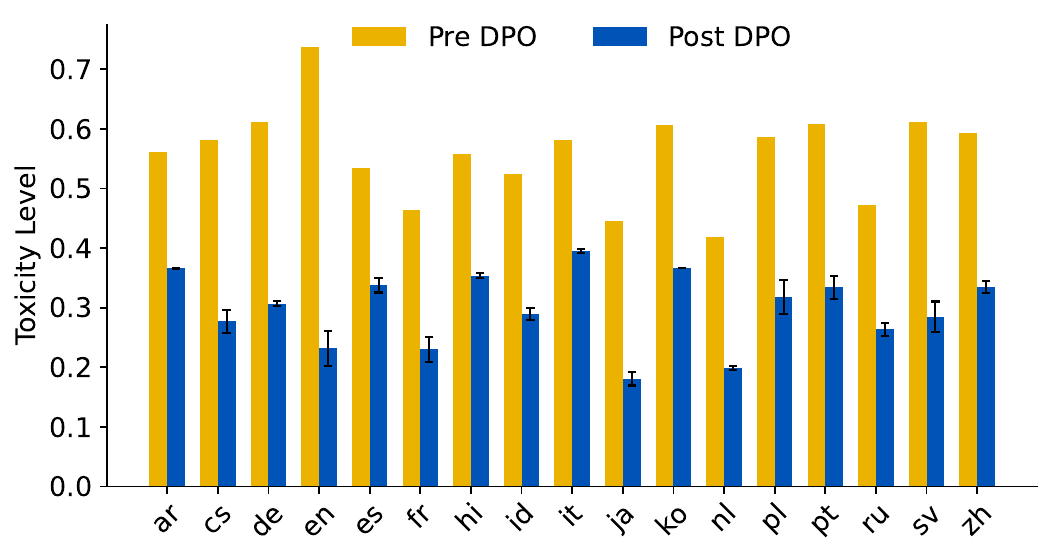}
         \caption{Expected maximum toxicity}
     \end{subfigure}
        \caption{Toxicity reduction of Aya-23 \citep{aryabumi2024aya23} after DPO training.}
        \label{fig:aya_dpo_result}
\end{figure*}

\section{Training Details}
\subsection{DPO Preference Tuning}
\label{app:dpo-hyperparam}

\begin{table}[htbp]
\centering
    \begin{tabular}{cc}
    \toprule
        Hyperparameter & Value\\
    \midrule
        Optimizer & RMSProp \\
        Learning Rate & 1E-5 \\
        Batch Size & 4 \\
        Gradient accumulation steps & 1 \\
        Loss & BCELoss \\
        Max gradient norm & 10 \\
        Validation metric & Loss/valid \\
        Validation patience & 10 \\
        DPO beta & 0.1 \\
        Epochs & 5 \\
    \bottomrule
    \end{tabular}
    \caption{Hyperparameters for DPO preference tuning for mGPT and BLOOM (1.7B).}
    \label{tab:dpo-hyperparam}
\end{table}

\subsection{Probe Training}
\label{app:probe-param}

\begin{table}[htbp]
\centering
    \begin{tabular}{cc}
    \toprule
        Hyperparameter & Value\\
    \midrule
        Optimizer & Adam\\
        Learning Rate & 0.0001\\
        Batch Size & 10\\
        Loss & BCELoss\\
        Epoch & 20 \\
    \bottomrule
    \end{tabular}
    \caption{Training hyperparameters for the binary toxicity classification probe $w_{\mathrm{toxic}}$.}
    \label{tab:probe-hyperparam}
\end{table}

\begin{figure*}
     \centering
     \begin{subfigure}[b]{0.48\textwidth}
         \centering
         \includegraphics[width=\textwidth]{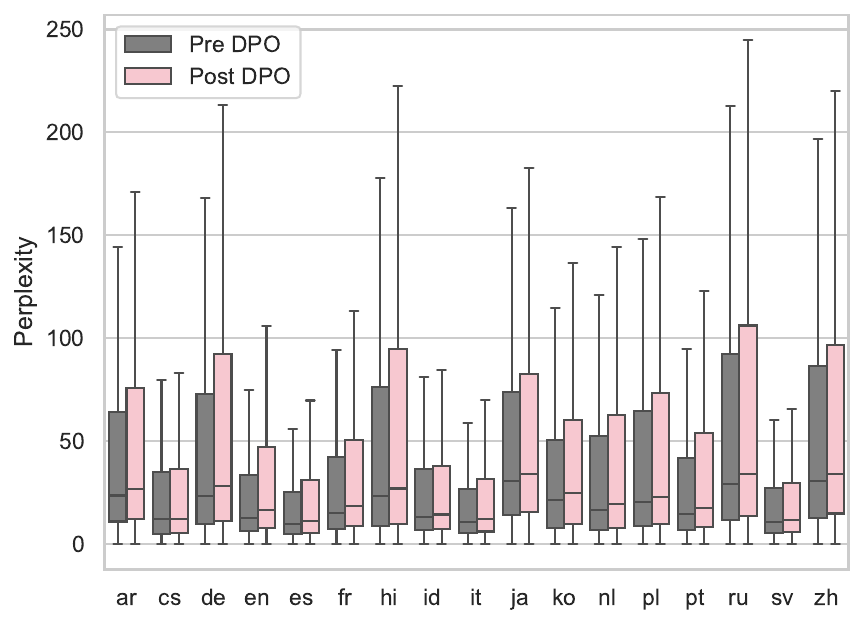}
         \caption{Box plot distribution of mGPT perplexity scores}
         \label{fig:dist_perplexity_scores_boxplot}
     \end{subfigure}
     \hfill
     \begin{subfigure}[b]{0.48\textwidth}
         \centering
         \includegraphics[width=\textwidth]{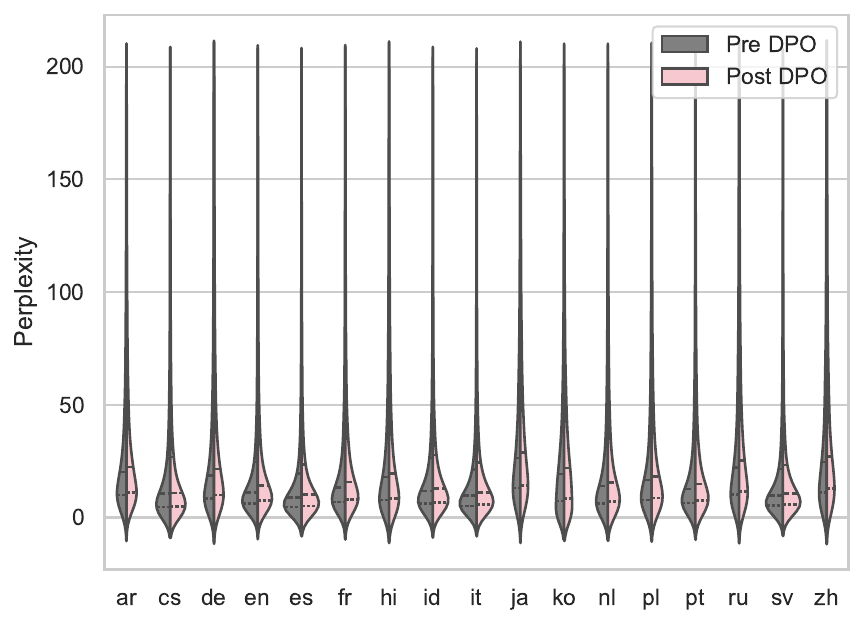}
         \caption{Violin plot distribution of mGPT perplexity scores}
     \end{subfigure}
        \caption{Per-language perplexity distribution of mGPT continuations before and after DPO training.}
        \label{fig:dist_perplexity_scores_violin}
        \label{fig:dist_perplexity_scores}
\end{figure*}

\section{Distribution of Perplexity Scores}
\label{app:distri-ppl-scores}
\Cref{fig:dist_perplexity_scores} displays the mGPT's distribution of the perplexity scores (which measures fluency) across all 17 languages. We observe that first, DPO preference tuning increases the perplexity of the generations as the median, interquatile range and whiskers increase in \Cref{fig:dist_perplexity_scores_boxplot}. Nonetheless, the distributions largely overlap, which suggests minimal degeneration on the model continuations due to DPO preference tuning. Second, the distributions in \Cref{fig:dist_perplexity_scores_violin} concentrate on reasonable range between 10 and 30 across different languages, and there are many outlier instances that leads to long tail distributions. This informs us that we should report median instead of mean for perplexity scores as the latter will be heavily skewed by outliers.

\begin{figure}
    \centering
    \includegraphics[width=.45\textwidth]{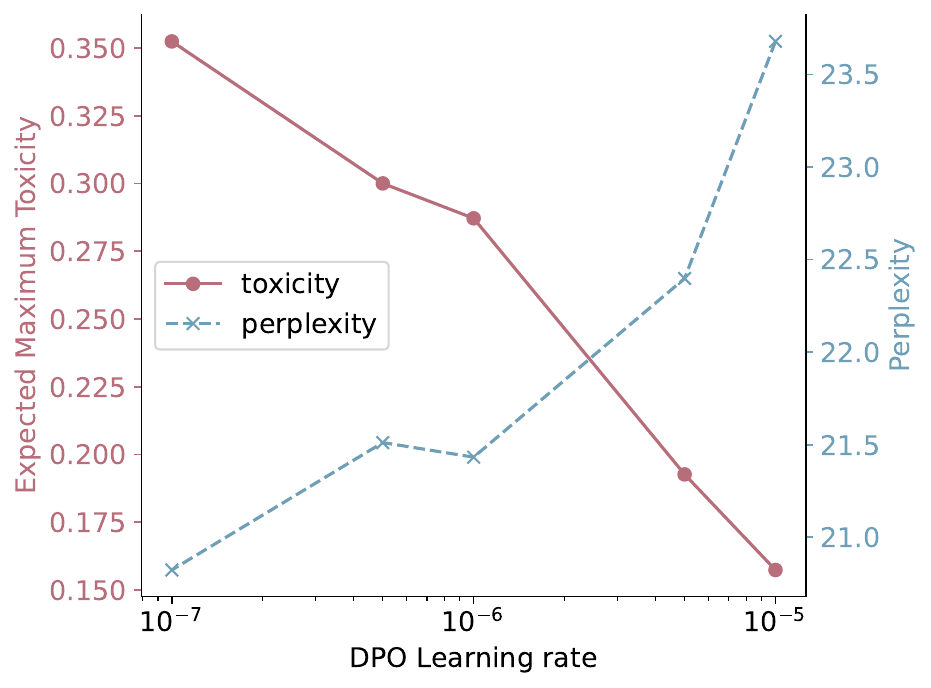}
    \caption{Tradeoffs between DPO learning rate, toxicity in post-DPO generation and perplexity across 17 languages.}
    \label{fig:tradeoff_lr_tox_ppl}
\end{figure}

\section{Tradeoffs between Learning Rate, Toxicity, and Perplexity Scores}
\label{app:tradeoffs-tox-ppl}
We perform English DPO training on mGPT model using the following five learning rate: \{1e-7, 5e-7, 1e-6, 5e-6, 1e-5\}, and we measure the toxicity level and fluency (perplexity) in model generations across 17 languages afterward. \Cref{fig:tradeoff_lr_tox_ppl} demonstrates the tradeoff between toxicity reduction and perplexity. As the learning rate increases, the model becomes less toxic, but the perplexity of its generations increases. We believe the reason is that since the RTP-LX input prompts are already contextually toxic, in which around 40\% of the prompts contain toxic words \citep{de2024rtplx}, generations that continue the \textit{toxic context} tends to be more natural than deliberating switching away from context for non-toxic continuations. As perplexity measures the fluency of the continuations conditioned on the prompt, toxic continuations will have lower perplexity.

\section{QLoRA and Multilingual Toxicity Reduction}
\label{app:qlora-toxicity-red}

\begin{figure}
    \centering
    \includegraphics[width=.49\textwidth]{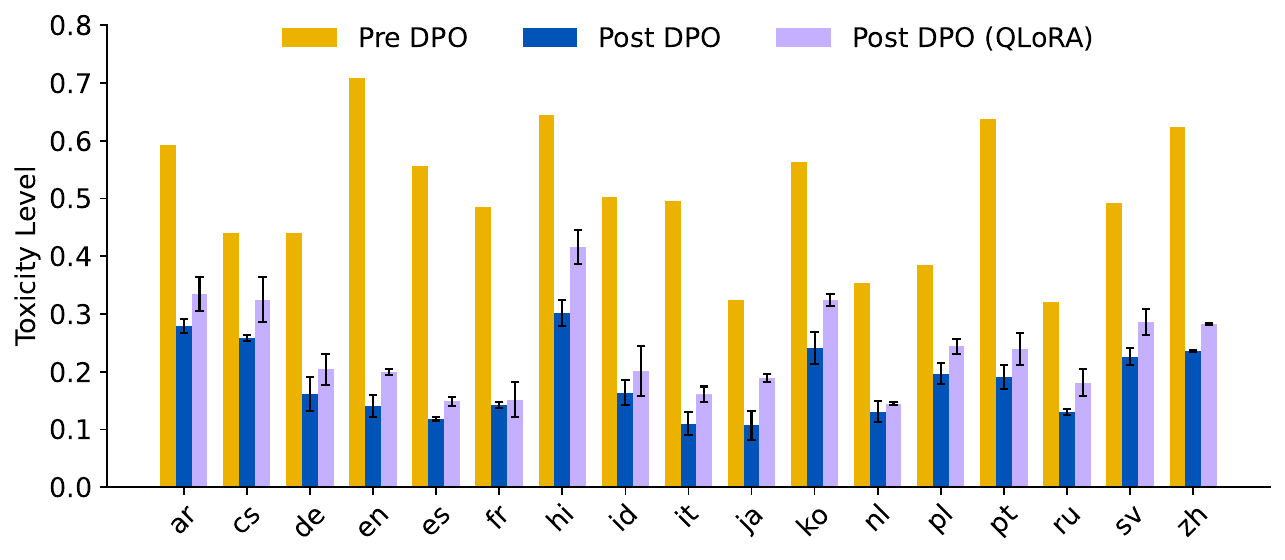}
    \caption{Comparison between full model training and QLoRA finetuning of BLOOM-1.7B with English DPO preference tuning.}
    \label{fig:qlora-bloom1b7}
\end{figure}

We perform full model finetuning and QLoRA finetuning of BLOOM-1.7B model with the same training hyperparameters in \Cref{tab:dpo-hyperparam} with the same number of training steps (up to convergence in 5-epoch training). \Cref{fig:qlora-bloom1b7} shows that model finetuned with QLoRA adapters remain more toxic than the full model finetuning. We believe this is due to QLoRA adapter finetuning has significantly less number of trainable parameters for same number of training steps.

\section{Bilingual Sentence Retrieval Experiment for Other LLMs}
\label{app:bloom-bilingual-retrieval}

\begin{figure}
    \centering
    \includegraphics[width=.45\textwidth]{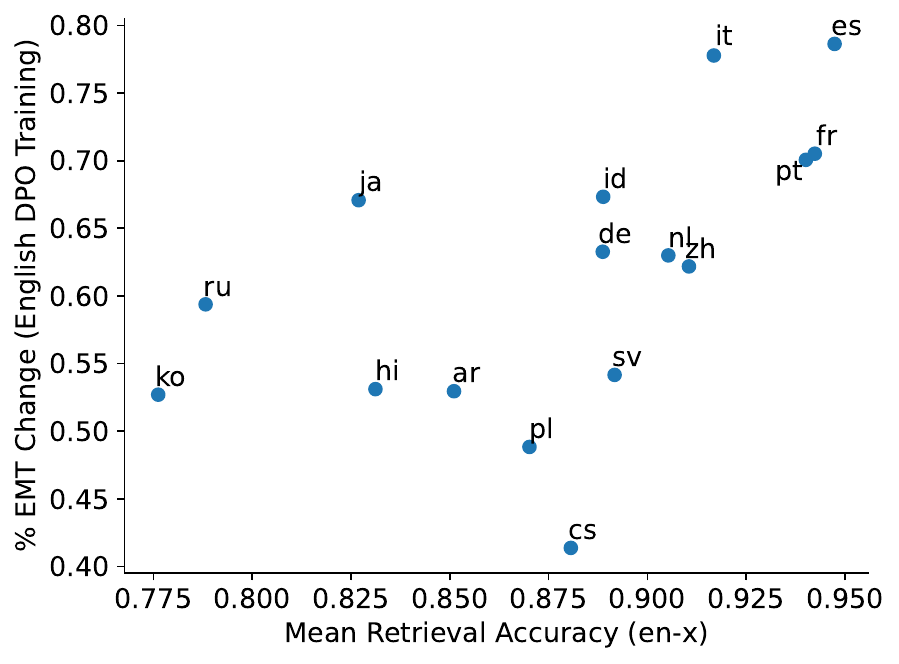}
    \caption{Percentage change in expected maximum toxicity against bilingual text retrieval accuracy for BLOOM-1.7B. Correlation with Pearson-r value of 0.59 (p < 0.01)}
    \label{fig:toxicity-retrieval-acc-bloom1b7}
\end{figure}

\begin{figure}
    \centering
    \includegraphics[width=.45\textwidth]{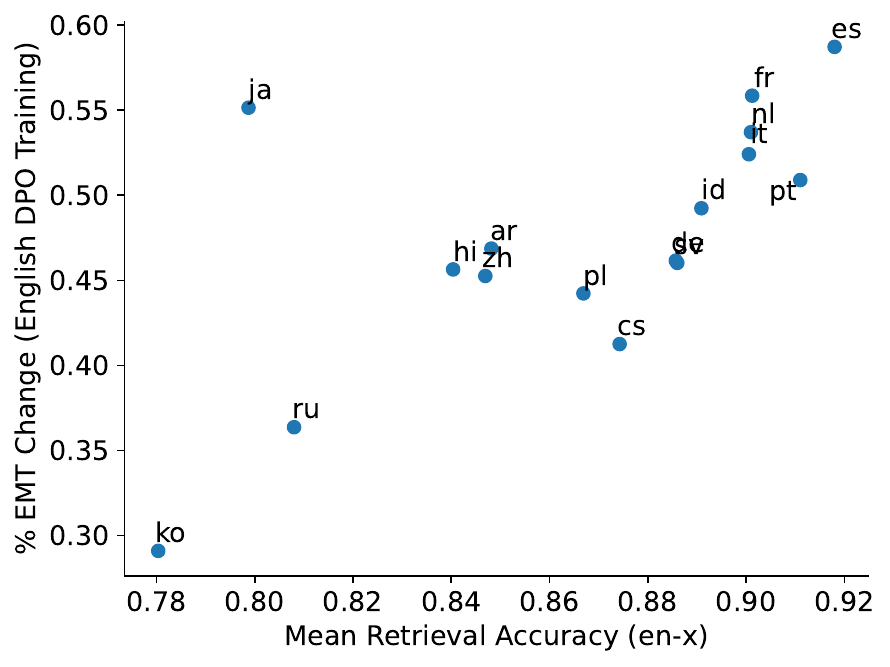}
    \caption{Percentage change in expected maximum toxicity against bilingual text retrieval accuracy for BLOOM-7.1B. Correlation with Pearson-r value of 0.66 (p < 0.01)}
    \label{fig:toxicity-retrieval-acc-bloom7b1}
\end{figure}

\begin{figure}
    \centering
    \includegraphics[width=.45\textwidth]{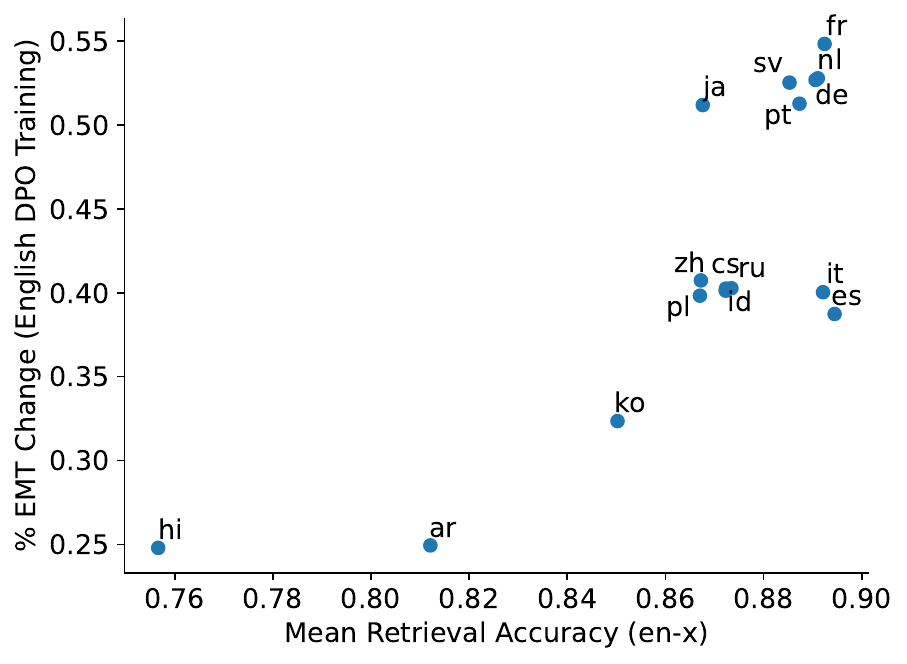}
    \caption{Percentage change in expected maximum toxicity against bilingual text retrieval accuracy for Llama2-7B. Correlation with Pearson-r value of 0.78 (p < 0.01)}
    \label{fig:toxicity-retrieval-acc-llama2}
\end{figure}

\Cref{fig:toxicity-retrieval-acc-bloom1b7}, \Cref{fig:toxicity-retrieval-acc-bloom7b1} and \Cref{fig:toxicity-retrieval-acc-llama2} show the positive correlation between bilingual sentence retrieval accuracy and percentage drop in EMT after English DPO training for BLOOM-1.7B, BLOOM-7.1B and Llama2-7B respectively. We observe similar findings as mGPT in \Cref{fig:toxicity-retrieval-acc}. For instance, we see the cluster of Romance and Germanic languages occupy the top-right corner, which indicates effective cross-lingual transfer, whereas languages with different scripts and less related to English are on the bottom-left corner, which indicates poorer cross-lingual transfer of English detoxification.

\section{Complete Table of Toxic Value Vectors}
\label{app:complete-table-neurons}
\Cref{tab:projection} presents the subset of value vectors identified as \textit{actual sources of toxicity}. 
For a comprehensive view, \Cref{tab:all-projection-1} and \Cref{tab:all-projection-2} include the complete list of all 36 vectors along with their projections.
Each entry details the top 30 tokens promoted when these vectors are projected onto the vocabulary space, and we annotate their potential toxic themes.
For clarity, the leading space is removed. 
Vectors are ranked according to their cosine similarities with the toxic probe vector $w_{\mathrm{toxic}}$. 
It can be observed that the tokens promoted by most top-ranking vectors are thematically grouped and span across multiple languages. For example, $w^3_{\mathrm{down},5794}$ promotes tokens related to pornography---in addition to common English tokens like ``porn'' and ``sex,'' it includes ``seks'' (sex in Malay), ``\includegraphics[width=0.039\textwidth]{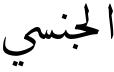}'' (sexual in Arabic), ``Член'' (a slang term in Russian meaning 'dick'), and ``\includegraphics[width=0.033\textwidth]{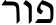}'' (a prefix in Hebrew equivalent to `por' in `porn'). While some tokens may not be inherently toxic, these projections clearly demonstrate the multilingual nature of the \textit{value vectors}.

\begin{table*}[htbp]
\small
  \centering
    \begin{tabular}{p{0.1\textwidth}p{0.17\textwidth}p{0.66\textwidth}}
        \toprule
        \textsc{Vectors} &\textsc{Toxic Theme} & \textsc{Promoted Tokens}\\
        \midrule 
        
        $w^{13}_{\mathrm{down}, 2337}$ 
            &Propaganda 
            & incomp, pseudo, manipul, propaganda, псев, ngu, corrupt, ignor, propagand, Propaganda, corrup, dece, manip, bankrupt, mercen, conspiracy, prét, conspira, fraud, blam, crimin, insult, selves, Emper, incap, пропаг, ignor, politiker, Politiker, massac\\
        \midrule
        $w^{14}_{\mathrm{down}, 6878}$ 
            &Exclamations
            & aa, ee, uu, EEE, aha, haa, mouth, hah, ah, oo, akka, pile, yy, !!, umph, öh, ее, wah, UU, моз, loo, Oh, ...)., oho, jee, oh, yah, ...], яя \\
        \midrule
        $w^{14}_{\mathrm{down}, 5723}$ 
            &Sexual Content 
            & sex, Sex, sex, porn, erot, Sex, seks, sexo, mast, Sexual, lesbian, anal, mature, sexual, сексу, Amateur, penetr, XXX, hardcore, sexuelle, Anal, blow, {đị}, amateur, domination, \includegraphics[width=0.03\textwidth]{main/singlewords/arabic1.pdf} \solvebad{جنس}, penet, osexual, sessuale, homosex\\
        \midrule
        $w^{9}_{\mathrm{down}, 6517}$ 
            &Negative Sentiments 
            & ngu, вообще, pula, мало, onaa, skull, alka, sick, poor, 졸, examp, yl, garr, meant, pessoal, dout, poik, \includegraphics[width=0.05\textwidth]{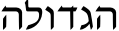} \solvebad{הגדולה}, 喪, feit, panas, liksom, cay, \includegraphics[width=0.044\textwidth]{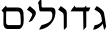}\solvebad{גדולים}, anat, \includegraphics[width=0.03\textwidth]{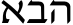}\solvebad{הבא}, tais, blam, önem, нуж\\
        \midrule
        $w^{12}_{\mathrm{down}, 6538}$ 
            &N/A
            & dreams, чуд, love, passion, Stories, \includegraphics[width=0.025\textwidth]{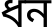}\solvebad{ধন}, fantas, magic, magia, heroes, !, битвы, venture, conquered, 昧, любви, wings, tough, fate, holy, хит, Geschichten, !』, thrown, ouro, cuore, фанта, !, storie\\
        \midrule
        $w^{12}_{\mathrm{down}, 6639}$ 
            &Gangster Culture 
            & maf, baller, Mafia, Hollywood, Cowboys, Buddy, Rocco, Maf, Cuban, Lords, rebell, istol, venes, bond, Johnny, Papa, Bobby, bourgeois, father, Boys, brothers, Wilde, latino, Dick, Buff, ègre, lesbian, Steele, rolle, latinos\\
        \midrule
        $w^{3}_{\mathrm{down}, 5794}$ 
            &Sexual Content 
            & porn, lesbian, sex, gay, sex, seks, sexo, сексу, \includegraphics[width=0.032\textwidth]{main/singlewords/arabic2-app.pdf}\solvebad{الجنسي}, Amateur, domination, xual, uremberg, Sex, Sex, Gay, escort, rape, amateur, hardcore, Gang, male, Член, Tub, XXX, girls, gang, \includegraphics[width=0.032\textwidth]{main/singlewords/hebrew15-app.pdf}\solvebad{פור}, hard, mass\\
        \midrule
        $w^{13}_{\mathrm{down}, 3368}$ 
            &N/A
            & EEE, unno, kee, kaa, Oh, おお, OK, !, uh, !』, Sweet, cierto, amn, tty, Girl, ohan, onaa, eh, ee, 何, Ok, ahu, Oh, 하자, knows, ok, \includegraphics[width=0.026\textwidth]{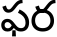}\solvebad{ఫర}, OK, ミー, Ok\\
        \midrule
        $w^{1}_{\mathrm{down}, 2583}$ 
            &Sexual Content 
            & sex, porn, lesbian, gay, sexo, сексу, Sex, \includegraphics[width=0.034\textwidth]{main/singlewords/arabic2-app.pdf} \solvebad{الجنسي}, seks, Sex, hardcore, rape, escort, Gay, sex, domination, Amateur, girls, celebrit, latina, ексу, mature, erot, revenge, Sexual, Girls, videos, sexuelle, \includegraphics[width=0.032\textwidth]{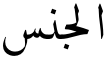}\solvebad{ الجنس}, tube\\
        \midrule
        ${w^{13}_{\mathrm{down},7176}}$
            &Sexual Content
            &sex, femenino, Femen, {сексу}, weib, girl, feminino, girls, Geschlechts, femen, Girls, {девуш}, women, sexo, Sex, Sexual, femmes, vrouwen, {γυνα}, Female, weibliche, {ексу}, féminine, féminin, femenina, Woman, Sex, femminile, {kvinnor}, {женщин} \\ 
        \midrule
        $w^{23}_{\mathrm{down}, 5888}$ 
        &N/A
        & K, S, D, H, Y, Y, F, W, R, N, T, P, K, G, DA, YA, YP, G, Z, M, H, IG, TAN, W, KS, S, O, E, IS, D\\
        \midrule
        $w^{8}_{\mathrm{down}, 7612}$ 
            &Severity and Crisis 
            & sév, 严重, èlement, fäll, icism, {loạn}, rophe, 嚴重, minaccia, endemic, に陥, menace, gravemente, akibat, amenaza, alkod, interference, interfer, szenved, caused, \includegraphics[width=0.02\textwidth]{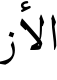}\solvebad{الأز}, spowod, {êne}, 壞, infolge, \includegraphics[width=0.033\textwidth]{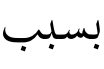}\solvebad{بسبب}, \foreignlanguage{vietnamese}{nặng}, \includegraphics[width=0.03\textwidth]{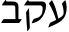}\solvebad{עקב}, sufr, enemigo\\
        \midrule
        $w^{11}_{\mathrm{down}, 7033}$ 
            &Counterculture 
            & funk, Evil, drummer, Chaos, Vampire, Punk, punk, Wrestling, Rotten, punk, Guns, Cody, Ghost, arious, Comedy, Superman, Teen, Hulk, ego, Theory, Kid, Funk, テレビアニメ, Girls, Mania, Johnny, Bee, Pokémon, girl, Hole\\
        \midrule
        $w^{11}_{\mathrm{down}, 4277}$ 
            & N/A
            & トップ, yard, floors, floor, publicly, кур, lap, Wet, пара, blow, рекор, open, back, Twitter, Sub, eplay, Live, オープ, boca, fermé, θμό, cean, pping, mouth, swing, **, пара, 閣, foot\\
        \midrule
        $w^{18}_{\mathrm{down}, 486}$ 
            &Destruction
            & saque, confisc, захват, \includegraphics[width=0.02\textwidth]{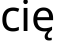}, assé, occupation, tho, 奪, ruin, cannon, 焼, gado, Пок, прода, vand, sell, \includegraphics[width=0.046\textwidth]{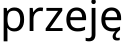}\solvebad{przeję}, \foreignlanguage{vietnamese}{chiếm}, аром, bezit, vine, devol, vand, conquest, verkocht, liqu, okup, \includegraphics[width=0.03\textwidth]{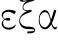}\solvebad{εξα}, burned, détr\\
        \midrule
        $w^{16}_{\mathrm{down}, 3531}$ 
            &Misconduct 
            & insult, abus, corrupt, prejud, fraud, confus, corruption, confusion, irrit, mauvais, 误, deform, scandal, {hại}, disastro, \begin{CJK*}{UTF8}{gbsn}{严重}\end{CJK*}, excessive, auvaise, 嚴重, disturb, abuse, violent, violations, degener, violation, corrup, poison, illeg, inad, irreg\\
        \midrule
        $w^{17}_{\mathrm{down}, 520}$ 
            &N/A
            & atson, oya, Lyc, arab, eldj, adino, arrista, arra, adin, arum, unak, ingles, ocha, Gall, rox, pup, olio, xen, ucia, arin, rina, utch, mala, wound, avin, arba, ellina, roa, oshi, cean\\
        \bottomrule
    \end{tabular}
  \caption{Projections of all 36 \textit{value vectors} from the \textit{actual sources of toxicity} - Part 1}
\label{tab:all-projection-1}
\end{table*}

\begin{table*}[htbp]
\small
  \centering
    \begin{tabular}{p{0.1\textwidth}p{0.17\textwidth}p{0.66\textwidth}}
        \toprule
        \textsc{Vectors} &\textsc{Toxic Theme} & \textsc{Promoted Tokens}\\
        \midrule
        $w^{12}_{\mathrm{down}, 3431}$ 
            &N/A
            & żon, heiratete, født, wander, がいる, whom, fri, married, ηλικ, geboren, 一人, elected, verheiratet, who, menik, naim, murdered, pope, diagnosed, convicted, heirat, casado, apell, candid, born, 晋, who, homeless, ermordet, resigned\\
        \midrule
        $w^{5}_{\mathrm{down}, 53}$ 
            &N/A
            & ilit, itre, egas, itur, íp, imet, utt, iag, ovi, urn, ocl, iny, orr, uttu, itab, imed, ipul, umed, iesa, udni, itore, igl, ittel, adah, enta, enn, ent, ierd, ulin, omm\\
        \midrule 
        $w^{10}_{\mathrm{down}, 4641}$ 
            &Exclamations
            & !", !, !".', !», !", !, !』, !!, ?", ！, !, ?".', orrow, ?", "»., \includegraphics[width=0.01\textwidth]{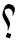}\solvebad{؟}, Who, "», +,, uu, "», Why, Your, survive, why, EEE, о�, ’", Tomorrow, …\\
        \midrule
        $w^{3}_{\mathrm{down}, 3173}$ 
            &Political Controversy 
            & \includegraphics[width=0.01\textwidth]{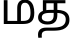}\solvebad{மத}, insult, criticism, accusations, allegations, Satan, polem, antisemit, boyc, Obama, attent, politician, gender, 념, atar, 罪, iste, ists, 民族, scandal, \includegraphics[width=0.033\textwidth]{main/singlewords/georgian1-tab3.pdf}\solvebad{აობ}, 支持, Massa, politically, Marl, Terror, contrad, istes, allegedly, uga\\
        \midrule
        $w^{7}_{\mathrm{down}, 3971}$ 
            &N/A
            & estri, ernas, erna, ater, īn, skapet, atine, ernes, atet, eza, enin, uses, isierte, riere, ija, ini, owskie, olin, ored, jekte, \includegraphics[width=0.024\textwidth]{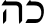}\solvebad{כה}, itty, okset, itäten, atina, liches, ated, \includegraphics[width=0.033\textwidth]{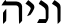}\solvebad{וניה}, ūr, izable\\
        \midrule
        $w^{16}_{\mathrm{down}, 4702}$ 
            &Struggle 
            & steen, sł, helping, 哨, shut, \includegraphics[width=0.027\textwidth]{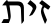}\solvebad{ זית}, elfen, стан, ума, \includegraphics[width=0.033\textwidth]{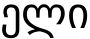}\solvebad{ელი}, 従, yal, \includegraphics[width=0.02\textwidth]{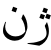}\solvebad{ژن}, hung, \includegraphics[width=0.036\textwidth]{main/singlewords/georgian1-tab3.pdf}\solvebad{აოდ}, 訓, 出力, help, ragon, embar, \includegraphics[width=0.027\textwidth]{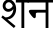}\solvebad{शन}, чер, ole, lopp, пех, пан, opes, \includegraphics[width=0.02\textwidth]{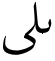}\solvebad{ یلی,} Олимпий, lief\\
        \midrule
        $w^{17}_{\mathrm{down}, 2392}$ 
            & N/A
            &\includegraphics[width=0.03\textwidth]{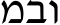}\solvebad{ובמ}, kad, eus, Са, \includegraphics[width=0.03\textwidth]{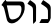}\solvebad{נוס}, kopol, wau, РО, uskan, kehr, holt, endien, YS, оган, aeus, жед, Дем, Hor, Пор, uais, uah, gekehr, rape, пед, onaa, sein, hold, Он, üsten\\
        \midrule 
        $w^{16}_{\mathrm{down}, 4689}$ 
            &Crime
            & destroyed, poison, broken, viola, \includegraphics[width=0.028\textwidth]{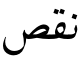}\solvebad{نقص}, violation, wrong, murdered, failure, destruction, траг, viol, 破, violations, killed, incendi, erro, suic, overth, obsol, \includegraphics[width=0.025\textwidth]{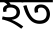}\solvebad{হত}, détruit, missing, Unf, faill, поврежд, uszkod, def, danneggi, fail\\
        \midrule
        $w^{21}_{\mathrm{down}, 7155}$ 
            &Geopolitical Tensions 
            & ίκης, リング, осс, ик, прек, ьют, \includegraphics[width=0.066\textwidth]{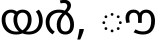}\solvebad{ യർ, ൗ}, ити, юс, енгер, бит, ета, еран, ерусалим, ̆, олот, болгар, АТУУ, ИК, СМИ, ън, уча, американских, ек, ък, \includegraphics[width=0.03\textwidth]{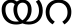}\solvebad{യറ}, французского, еке\\
        \midrule
        $w^{0}_{\mathrm{down}, 7248}$ 
        & N/A
        & ンプ, унь, udad, bett, \includegraphics[width=0.036\textwidth]{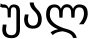}\solvebad{უალ}, стри, \includegraphics[width=0.026\textwidth]{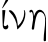}\solvebad{ίνη}, \includegraphics[width=0.036\textwidth]{main/singlewords/georgian1-tab3.pdf}\solvebad{აობ}, мини, orki, \includegraphics[width=0.028\textwidth]{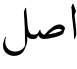}\solvebad{اصل}, Mandat, ziali, Pict, orsi, Bata, 渉, sculpt, ма, partij, осто, орот, inea, marker, Massa, \includegraphics[width=0.032\textwidth]{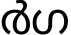}\solvebad{ർഗ}, Pem, inten\\
        \midrule
        $w^{17}_{\mathrm{down}, 3530}$ 
            &N/A
            & СА, DE, КО, DO, OF, DA, TO, THE, TE, DOS, CA, TH, SI, NA, WA, SH, DI, RE, БА, LA, PA, AN, ME, SO, TU, OR, MA, FL, EN, ВС\\
        \midrule
        $w^{23}_{\mathrm{down}, 2675}$ 
            &Legislation Terms 
            & 抵, 本身, \includegraphics[width=0.026\textwidth]{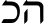}\solvebad{הכ}, 。（, 見られる, importante, 一般的, みられる, 人で, 。」, essoort, ，“, menoptera, 。”, 交代, тины, などがある, \includegraphics[width=0.041\textwidth]{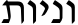}\solvebad{וניות},
            \includegraphics[width=0.041\textwidth]{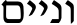}\solvebad{וניים}, 建制, 。《, 最多, 可能是, \includegraphics[width=0.030\textwidth]{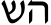}\solvebad{הש}, 色的, こく, 特的, 法的, 划, 学名\\
        \midrule
        $w^{11}_{\mathrm{down}, 3027}$ 
            &N/A 
            & OK, \includegraphics[width=0.030\textwidth]{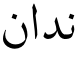}\solvebad{ندان}, cinese, ̆, hide, ену, хе, jade, 撲, \includegraphics[width=0.030\textwidth]{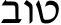}\solvebad{טוב}, sea, boys, afterwards, chines, ковой, broke, hung, енский, лё, rocks, endem, normal, ть, quit, 二世, europé, otherwise, Москва, allemande, bourg\\
        \midrule
        $w^{10}_{\mathrm{down}, 8010}$ 
            &Sexual Content 
            &\includegraphics[width=0.033\textwidth]{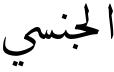}\solvebad{الجنسي}, couples, sex, Geschlechts, femen, 貞, lesbian, seks, Sex, sex, kontrak, seksual, femenina, Sex, feminist, sexual, Femen, masculino, 育, 合意, mulheres, женщин, women, féminin, nat, secondaires, femenino, женат, \includegraphics[width=0.033\textwidth]{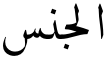}\solvebad{الجنس}\\
        \midrule
        $w^{10}_{\mathrm{down}, 2127}$ 
            & N/A
            & e, ament, es, en, edades, enes, idades, ues, eni, ате, ив, ería, ute, ений, ibles, \includegraphics[width=0.028\textwidth]{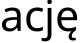}\solvebad{ację}, ere, ата, enie, entes, ente, ate, uos, ió, ies, ения, ables, eniu, osos, esi\\
        \midrule
        $w^{15}_{\mathrm{down}, 594}$ 
            & N/A
            & ER, EN, ING, AST, ERS, DE, OF, ENT, LAN, EL, UL, THE, IN, TAT, EM, OR, ASS, LO, ET, YA, HE, ON, AN, ISE, CON, IST, CH, INE, DO, RO\\
        \midrule
        $w^{10}_{\mathrm{down}, 7751}$ 
            & N/A
            & rol, stein, \includegraphics[width=0.028\textwidth]{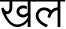}\solvebad{खल}, uba, dic, romos, ecin, dül, deling, icip, \includegraphics[width=0.032\textwidth]{main/singlewords/georgian3-app.pdf}\solvebad{δή}, duk, {stä}, \includegraphics[width=0.022\textwidth]{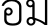}\solvebad{อม}, sor, veen, kül, tuk, band, 克斯, upe, ahnya, {gång}, ysis, scy, ragalus, зен, dem, {föd}, \includegraphics[width=0.023\textwidth]{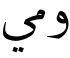}\solvebad{ومي}\\

        \midrule
        $w^{10}_{\mathrm{down}, 4920}$ 
            & N/A
            & British, hemp, bull, badan, Billie, rump, BB, dada, berkembang, rien, gede, berupa, sph, woman, Ку, ео, Sub, dik, uber, Traff, худ, tartott, boca, Britain, fell, discográfica, brutal, Mel, bong, allow\\
        \midrule
        $w^{14}_{\mathrm{down}, 7052}$ 
        & Severe Condition
        & атастро, failure, violation, insult, disastro, 误, катастро, потери, 陷, catast, потеря, disturb, неуда, 失, deterior, 嚴重, \includegraphics[width=0.03\textwidth]{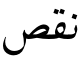}\solvebad{نقص}, violations, 危, \begin{CJK*}{UTF8}{gbsn}{严重}\end{CJK*}, confusion, disappoint, наруш, discont, 傷, 事故, 違反, worst, confus, conflict\\
    \bottomrule
    \end{tabular}
    \caption{Projections of all 36 \textit{value vectors} from the \textit{actual sources of toxicity} - Part 2}
    \label{tab:all-projection-2}
\end{table*}

\end{document}